\documentclass[sigconf,nonacm]{acmart}

\setcopyright{none}
\usepackage{amsmath,amsfonts}
\usepackage{algorithm}
\usepackage{algorithmic}
\usepackage{array}
\usepackage{booktabs}
\usepackage{multirow}
\usepackage{graphicx}
\usepackage{float}
\usepackage{tabularx}
\usepackage{placeins}
\usepackage[caption=false]{subfig}
\usepackage{textcomp}
\usepackage{url}
\usepackage{verbatim}

\newcommand{\method}{CLEAR}

\begin{document}

\title{\method: Coupling Local Evidence with Affective Representations for Multimodal Emotion Recognition in Conversation}


\author{%
\parbox{0.98\textwidth}{%
\centering
\resizebox{0.98\linewidth}{!}{%
\textbf{%
Linan Zhu$^{1}$ \quad
Zihao Zhai$^{1}$ \quad
Mengfei Xu$^{1}$ \quad
Xiao Han$^{1,*}$ \quad
Yuqian Fu$^{2}$ \quad
Xiangfan Chen$^{1}$ \quad
Zihang Zhang$^{1}$
}%
}%
}%
}

\renewcommand{\shortauthors}{Zhu et al.}

\begin{abstract}
In video calls and livestreamed discussions, recognizing speakers' emotions helps identify emotionally salient segments and supports fine-grained video content analysis.
Multimodal Emotion Recognition in Conversation (ERC) provides an approach to recognizing emotions in such content by integrating textual, visual, and acoustic information with conversational context.
Current methods largely leverage the contextual understanding capabilities of multimodal large language models (MLLMs) to predict emotions by generating labels.
However, emotion understanding and label prediction may be inconsistent, leaving the internal affective information underutilized.
Moreover, when the current utterance is jointly modeled with extensive dialogue history and multimodal information, its local discriminative cues may be diluted and become less salient in the resulting contextual representation.
To address these challenges, we propose \textbf{CLEAR}, a framework for \textbf{C}oupling \textbf{L}ocal \textbf{E}vidence with \textbf{A}ffective
\textbf{R}epresentations in ERC.
To mitigate the mismatch between emotion understanding and label prediction, we use the continuous internal representation of an emotion-supervised vision-language model (VLM) instead of its generated emotion labels, and further train a classifier based on this representation.
To strengthen local discriminative cues, we use sample-adaptive residual coupling to dynamically integrate textual and acoustic evidence from the target utterance into the contextual representation.
Experiments on IEMOCAP and MELD demonstrate that \method{} achieves competitive performance, supporting the effectiveness of the proposed approach.
The code is available at \url{https://anonymous.4open.science/r/vlm-210F/}.
\par\smallskip
\noindent\textbf{Relevance Statement:}
This paper investigates emotion recognition in multimodal online video conversations and livestreams.
The proposed method can facilitate the identification of emotionally salient segments in livestreamed content and support fine-grained online video content analysis.
\end{abstract}

\maketitle

\begingroup
\renewcommand{\thefootnote}{}
\footnotetext{%
$^{1}$ Zhejiang University of Technology, Hangzhou, China;
$^{2}$ ETH Zurich, Switzerland.
$^{*}$ Corresponding author: Xiao Han.
}
\endgroup

\section{Introduction}
\label{sec:introduction}

Emotion is an important cue for understanding users' intentions and behaviors.
In video calls and livestreamed discussions, users express emotions through spoken content, facial expressions, and vocal cues.
Recognizing these emotions can help identify emotionally salient video segments and support fine-grained analysis of online conversational video content.
Multimodal Emotion Recognition in Conversation (ERC) is a task closely related to this need, aiming to identify the emotional state of each utterance by jointly using textual, visual, and acoustic information.
Accurate recognition requires jointly exploiting the local expressive cues of the target utterance and the broader conversational context.
Local cues such as textual content, intonation, and pauses provide direct evidence for emotion discrimination.
In contrast, contextual information such as dialogue history and speaker interactions helps interpret what these cues actually mean~\cite{majumder2019dialoguernn,ghosal2019dialoguegcn}.
Early ERC methods fused multimodal features and gradually incorporated dialogue history, speaker interactions, and cross-modal relationships to obtain increasingly rich contextual information~\cite{hu2022mm,chudasama2022m2fnet,shi2023multiemo,yi2024haucl}.

\begin{figure}[t]
    \centering
    \includegraphics[width=\linewidth]{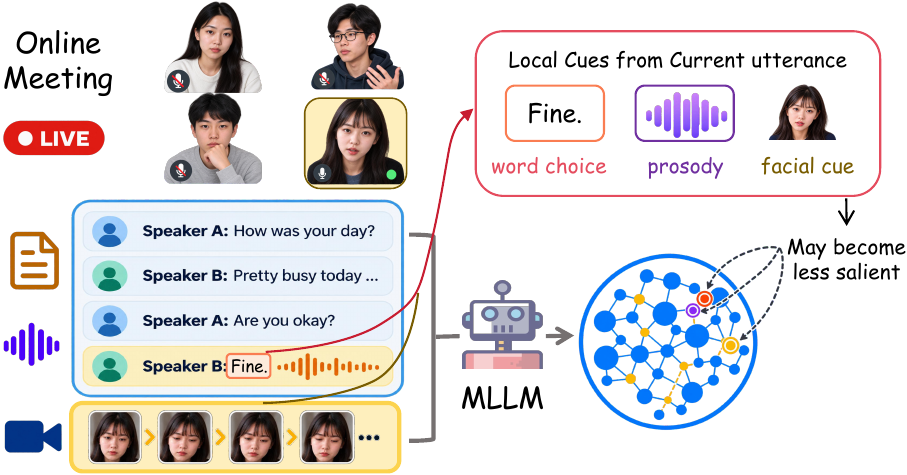}
    \caption{
        Conceptual illustration of potential local discriminative cue dilution in multimodal contextual modeling of an online meeting.
         Local cues from the target utterance, such as word choice, prosody, and facial expressions, may become less salient in the resulting rich contextual representation.
}
    \label{fig:motivation}
\end{figure}
These methods remain limited in understanding emotions in complex conversational contexts~\cite{hu2022mm,chudasama2022m2fnet,shi2023multiemo,yi2024haucl}.
Recently, multimodal large language models (MLLMs) have offered a new approach to ERC
owing to their strong contextual understanding.
Existing MLLM-based methods usually adapt these models through instruction tuning~\cite{lei2023instructerc,zhang2025dialoguellm} or parameter-efficient fine-tuning~\cite{chen-etal-2025-grpo}.
They then obtain the final prediction by generating emotion labels~\cite{lei2023instructerc,zhang2025dialoguellm,chen-etal-2025-grpo,wu2025beyond,he2025dialoguemmt,fu2025bemerc}.
However, studies such as MME-Emotion report an inconsistency between emotion understanding and label prediction in MLLMs~\cite{zhang2026mme}.
A model may correctly identify emotional cues yet generate an incorrect label, or generate a correct label based on inaccurate cues.
This indicates that the emotion understanding of MLLMs has not been fully translated into reliable emotion discrimination.
Relying solely on generated discrete labels may therefore fail to fully exploit the emotion-related information encoded inside the model.
This leads to \textbf{Challenge 1: how can the inconsistency between emotion understanding and discrimination in MLLMs be mitigated to better exploit their internal affective information?}

By jointly modeling the target utterance, dialogue history, and multimodal information, MLLMs can provide important support for emotion recognition.
However, prior work shows that large models do not use all information in a long context equally well~\cite{shi2023large,liu2024lost}.
In an online meeting, a target utterance may be as brief as ``Fine.'', surrounded by multi-turn dialogue, speaker visual cues, and acoustic information.
As illustrated in Fig.~\ref{fig:motivation}, these inputs are jointly modeled by an MLLM to form a multimodal contextual representation, where a small highlighted node schematically represents local discriminative information.
This illustration suggests that fine-grained cues such as intonation and pauses, despite being included in joint modeling, may become less salient in the resulting rich contextual representation.
This leads to \textbf{Challenge 2: how can the role of local discriminative cues be strengthened while the rich contextual information of MLLMs is retained?}

To address these two challenges, we propose \textbf{CLEAR},
a framework for \textbf{C}oupling \textbf{L}ocal
\textbf{E}vidence with \textbf{A}ffective
\textbf{R}epresentations in ERC, which integrates multimodal
contextual representations with local discriminative cues.
It consists of two complementary components: emotion-supervised VLM adaptation and sample-adaptive residual coupling.
For Challenge 1, we adapt the VLM through ERC-supervised LoRA fine-tuning to enhance its internal emotion-discriminative representations.
Rather than relying on generated emotion labels, we extract its continuous internal representation as a multimodal contextual representation.
We then train a classifier on this representation for emotion discrimination in conversations.
For Challenge 2, we introduce sample-adaptive residual coupling to strengthen local discriminative cues while preserving contextual information.
We additionally extract textual and raw acoustic features from the target utterance as local discriminative evidence.
Sample-adaptive residual fusion then dynamically adjusts their contributions to the contextual representation.

The main contributions of this work are summarized as follows:
\begin{itemize}
    \item We adapt a general-purpose VLM with ERC emotion supervision and further train a classifier using its continuous internal hidden representation, promoting alignment between emotion understanding and discrimination.
    \item We propose a sample-adaptive residual coupling mechanism that adaptively fuses the local textual and acoustic evidence of the target utterance with the contextual representation.
    \item We conduct systematic experiments on IEMOCAP and MELD, and ablation studies and further analyses verify the effectiveness of the proposed framework.
\end{itemize}
\section{Related Work}
\label{sec:related-work}
\subsection{Context Modeling and Multimodal Fusion}

Early research on emotion recognition in conversation (ERC) mainly focused on modeling conversational context, speaker interactions, and dependencies among utterances. For example, DialogueRNN~\cite{majumder2019dialoguernn} tracks participant states and conversational context to model emotional dynamics, DialogueGCN~\cite{ghosal2019dialoguegcn} uses graph structures to capture speaker-dependent contextual dependencies, and CoMPM~\cite{lee2022compm} tracks the pre-trained memory of each speaker.
With the development of multimodal affective analysis, ERC research turned to the complementary information across textual, acoustic, and visual modalities. Early studies explored joint representation learning across modalities: Aguilar et al.~\cite{aguilar2019multimodal} combined lexical and acoustic information, and Multilogue-Net~\cite{shenoy2020multilogue} extended multimodal modeling to conversational settings with contextual and participant-state information. MM-DFN~\cite{hu2022mm}, M2FNet~\cite{chudasama2022m2fnet}, and MultiEMO~\cite{shi2023multiemo} later improved multimodal modeling through dynamic fusion, dedicated cross-modal interaction, and correlation-aware attention, respectively. Adaptive graph learning has also been applied to multimodal conversational emotion detection~\cite{tu2024adagin}.
More recent studies address complex conversational relationships and finer-grained multimodal interactions: DER-GCN~\cite{ai2025dergcn} explores graph-based relational modeling, HAUCL~\cite{yi2024haucl} combines a hypergraph autoencoder with contrastive learning, and ECERC~\cite{zhang-tan-2025-ecerc} models the interaction between emotional evidence and conversational context. CSS~\cite{lyu2026css} and EMART~\cite{seong2026emart} study multimodal emotion modeling through cross-space synergy and audio-referred textual representations, respectively. Overall, these methods have moved from conversational context modeling toward more sophisticated multimodal interaction and fusion, but they generally still rely on specially designed contextual encoders and fusion architectures, and emotion understanding in complex conversational contexts remains limited.

\subsection{LLM/VLM-Based ERC}

Multimodal large models have been increasingly adopted for ERC owing to their strong contextual understanding and multimodal modeling abilities, especially for interpreting context-dependent emotions across conversational turns. InstructERC~\cite{lei2023instructerc} and DialogueLLM~\cite{zhang2025dialoguellm} formulate ERC as instruction- and context-aware emotion prediction through instruction tuning and emotion knowledge modeling, respectively, which allows LLMs to exploit the dialogue history for emotion judgment. To exploit non-textual cues, SpeechCueLLM~\cite{wu2025beyond} converts speech characteristics into natural language descriptions added to the prompts, whereas DialogueMMT~\cite{he2025dialoguemmt} and BeMERC~\cite{fu2025bemerc} process multimodal conversational information and speaker behavior more directly. Recent studies further explore optimization and reasoning strategies: GRPO-LLM~\cite{chen-etal-2025-grpo} focuses on modality selection and optimization, Causal-ERC~\cite{jing2026causal} introduces causal prompting to guide emotion reasoning, and PRC-Emo~\cite{li2026prcemo} combines prompt retrieval with curriculum learning to improve task adaptation.
Other lines of work also incorporate speaker characteristics into LLM-based ERC~\cite{fu2025laerc}, explore LLM-supervised pre-training for multimodal emotion representations~\cite{dutta2025llm}, and investigate cognitive modeling with traits and distortions for multimodal ERC~\cite{nguyen2026cmtd}. EmoMM further specifically evaluates multimodal emotion recognition under conditions involving conflicting and missing modality information~\cite{sun2026emomm}.

These studies have enhanced the use of dialogue history, speaker information, and multimodal cues in large models for ERC.
Dialogue history and speaker interactions help interpret emotional expressions more effectively in context, while textual, acoustic, and visual cues from the target utterance provide direct evidence for emotion discrimination.
In this work, we adapt a VLM with emotion supervision, extract its continuous hidden representation as a multimodal contextual representation, and train an additional classifier for emotion discrimination. We further use context-guided acoustic attention and sample-adaptive residual coupling to integrate local textual and acoustic evidence from the target utterance, dynamically adjusting their contributions.

\section{Method}
\label{sec:method}
\subsection{Overview}
\label{sec:overview}

\begin{figure*}[t]
    \centering
    \includegraphics[width=\textwidth]{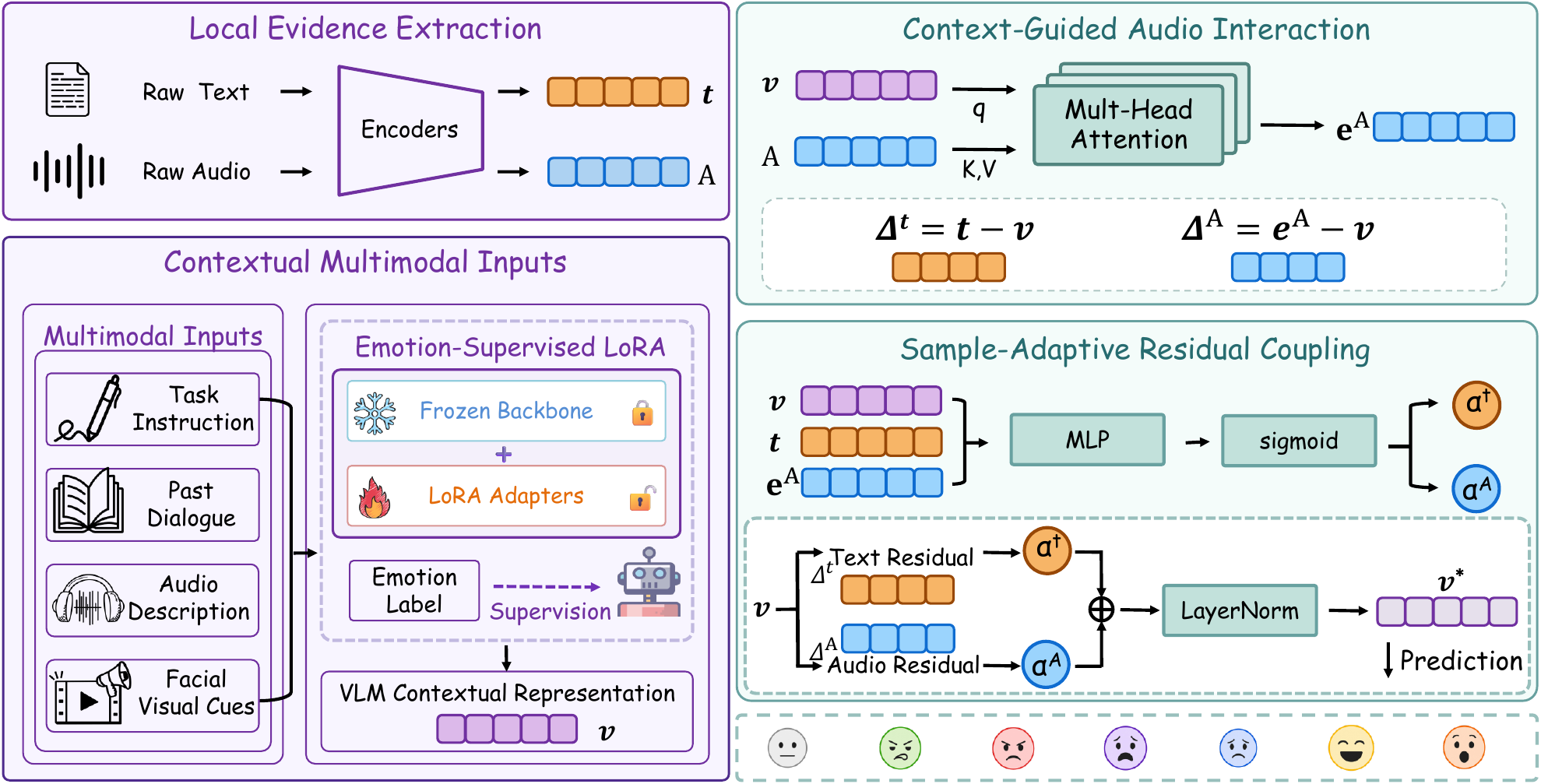}
    \caption{
        Overview of the proposed \method{} framework, including emotion-supervised contextual representation learning, local evidence extraction, context-guided audio interaction, and sample-adaptive residual coupling.
    }
    \label{fig:framework}
\end{figure*}

As illustrated in Fig.~\ref{fig:framework}, \method{} couples a contextual representation with local discriminative cues from the current utterance.
\method{} has two complementary pathways.
The first pathway obtains a continuous contextual representation $v_{\mathrm{raw}}$ from multimodal conversational inputs using an emotion-supervised VLM.
The second pathway uses dedicated feature extractors to obtain the textual feature $t_{\mathrm{raw}}$ and frame-level acoustic features $A_{\mathrm{raw}}$ of the target utterance.
After mapping these features into a shared space, the contextual representation $v$ guides acoustic attention, while a residual router adjusts the contributions of textual and acoustic residuals.
The resulting unified representation $v^*$ is used for emotion prediction, with additional supervision from contrastive learning and auxiliary classification during training.


\subsection{Contextual Representation Learning}
\label{sec:situational-representation}
Emotion understanding and label prediction may be inconsistent in VLMs, so relying solely on generated discrete labels may underutilize their internal affective information.
Continuous hidden representations can retain learned contextual information and provide richer representations for downstream emotion discrimination and local-evidence fusion.
We therefore learn a contextual representation through emotion-supervised VLM adaptation and retain its continuous hidden state for subsequent fusion.
Specifically, the multimodal input $X$ consists of the target utterance, target-speaker facial images, up to ten preceding utterances, and a high-level acoustic description generated by Qwen2-Audio-7B-Instruct.
For visual information, we first localize the target speaker using active speaker detection (ASD) and then uniformly sample four facial frames from the video segment corresponding to the target utterance.
For dialogue history, we take up to ten strictly preceding utterances with their speaker identifiers.
For acoustic information, we use Qwen2-Audio-7B-Instruct~\cite{chu2024qwen2audio} to generate a textual high-level description of the target audio.
The generation is conditioned on the target audio, the current transcript, the target-speaker identifier, and up to ten preceding utterances.
The prompt for audio-description generation is shown in Fig.~\ref{fig:prompt-design}(a) in Appendix~\ref{app:prompts}.

Given these multimodal inputs, we use Qwen3-VL-8B-Instruct~\cite{bai2025qwen3vl} as the VLM backbone and apply ERC-oriented emotion-supervised adaptation with LoRA~\cite{hu2022lora} to further elicit its emotion-discriminative capability.
For each dataset, we train a separate LoRA model using only its training split.
During adaptation, the pretrained VLM backbone remains frozen, and only the LoRA parameters are optimized to predict the ground-truth emotion label of the current target utterance from the multimodal conversational input.
The prompt used for emotion-supervised VLM adaptation is shown in Fig.~\ref{fig:prompt-design}(b) in Appendix~\ref{app:prompts}.

The adaptation loss is computed only over the ground-truth emotion label tokens:
\begin{equation}
\mathcal{L}_{\mathrm{adapt}}
=
-\sum_{j\in \mathcal{Y}}
\log
p_{\theta_0+\Delta\theta_{\mathrm{LoRA}}}
\left(y_j\mid X,y_{<j}\right),
\label{eq:adapt}
\end{equation}
where $\theta_0$ denotes the pretrained VLM parameters, $\Delta\theta_{\mathrm{LoRA}}$ denotes the trainable LoRA parameters, and $\mathcal{Y}$ denotes the index set of the ground-truth emotion label tokens.

Specifically, we take the hidden state at the last valid position of the input prompt in the final VLM layer:
\begin{equation}
v_{\mathrm{raw}}
=
H_{\theta_0+\Delta\theta_{\mathrm{LoRA}}}^{(L)}
(X)[p],
\qquad
v_{\mathrm{raw}}\in\mathbb{R}^{4096},
\label{eq:vlm-representation}
\end{equation}
where $L$ denotes the final VLM layer and $p$ denotes the last valid position of the input prompt, prior to emotion label generation.
Because this hidden state is conditioned on the complete multimodal input, it encodes contextual information about the current utterance, the dialogue history, the target-speaker visual information, and the high-level acoustic semantics.


\subsection{Utterance-Specific Cue Extraction}
\label{sec:local-cues}

Although the contextual representation integrates rich conversational information, local discriminative cues from the target utterance may become less salient during joint modeling.
Relying solely on this representation may therefore underutilize fine-grained textual and acoustic information from the target utterance.
To address this limitation, we use dedicated textual and acoustic encoders to extract utterance-specific discriminative evidence that complements the contextual representation.
For the textual modality, we use RoBERTa-large~\cite{liu2019roberta} to encode the current target utterance and take its CLS representation as the utterance-level textual feature, denoted by $t_{\mathrm{raw}}$.
For the acoustic modality, we use Emotion2vec-large~\cite{ma2024emotion2vec} to encode the complete speech signal of the current target utterance.
Rather than aggregating the acoustic output into a single utterance-level vector right away, we keep the complete frame-level acoustic feature sequence $A_{\mathrm{raw}}$ to preserve fine-grained temporal information such as prosodic variation and pauses, and use it for the later context-guided acoustic cue extraction.


\subsection{Sample-Adaptive Residual Coupling}
\label{sec:adaptive-coupling}

The contributions of local textual and acoustic evidence to emotion discrimination may vary across utterances, requiring sample-specific adjustment of local evidence.
To this end, we propose a sample-adaptive residual coupling mechanism that combines context-guided acoustic attention with a learnable residual router to dynamically integrate local discriminative evidence while retaining contextual information.
The previous steps yield three raw representations: the textual feature $t_{\mathrm{raw}}$, the acoustic feature sequence $A_{\mathrm{raw}}$, and the contextual representation $v_{\mathrm{raw}}$.
To support cross-modal interaction, we map the three representations into a shared 512-dimensional feature space.
For the textual modality, a linear projection followed by LayerNorm maps $t_{\mathrm{raw}}$ into the shared space, giving $t$.
For the acoustic modality, $A_{\mathrm{raw}}$ is first projected into the shared space and then processed by a trainable two-layer Transformer, resulting in the frame-level acoustic sequence $A$.
Attention pooling then gives the utterance-level acoustic representation $\bar{A}=\sum_{l=1}^{T}\pi_l A_l$, while the complete sequence $A$ is kept for the later cross-attention,
where $T$ is the number of valid acoustic frames, and $\pi_l$ is the Softmax-normalized score of $A_l$ given by a learnable scoring function.
For the contextual representation, $v_{\mathrm{raw}}$ is first projected into the shared 512-dimensional space by a linear layer and then refined by an MLP, which gives $v$.

Using the contextual representation $v$ as a reference, we introduce textual and acoustic local cues from the current utterance and couple them with the contextual representation through residual modeling.
For the textual modality, we use the textual representation $t$ directly as the local textual cue.
For the acoustic modality, we apply multi-head cross-attention to obtain context-guided local acoustic cues:
\begin{equation}
e^A
=
\operatorname{MHA}
\left(
Q=v,\,
K=A,\,
V=A
\right),
\label{eq:cross-attention}
\end{equation}
where $e^A$ denotes the context-guided local acoustic cue, with the contextual representation $v$ as the query and the frame-level acoustic sequence $A$ as the keys and values.

We use a learnable sample-adaptive gating network $G$ to estimate the contributions of textual and acoustic residuals relative to the contextual representation.
The weighted residuals are added to the contextual representation, followed by LayerNorm, which gives the fused representation:
\begin{equation}
v^*
=
\operatorname{LN}
\left(
v
+
\alpha^t\Delta^t
+
\alpha^A\Delta^A
\right),
\label{eq:adaptive-refinement}
\end{equation}
where $\Delta^t=t-v$ and $\Delta^A=e^A-v$ denote the textual and acoustic residuals relative to the contextual representation.
The gating coefficients $[\alpha^t,\alpha^A]=\sigma(G([v;t;e^A]))\in(0,1)^2$ are estimated by $G$, a two-layer MLP, from the concatenation of $v$, $t$, and $e^A$.

Finally, the fused representation $v^*$ is fed into an MLP classifier $f_{\mathrm{cls}}(\cdot)$, followed by Softmax to obtain the emotion-class probability distribution $\hat{\mathbf{p}}=\operatorname{Softmax}(f_{\mathrm{cls}}(v^*))$.

\subsection{Learning Objective}
\label{sec:learning-objective}

The final emotion classification loss alone does not explicitly enforce affective semantic consistency among the textual, acoustic, and contextual representations.
Moreover, it does not impose a separate emotion-discriminative constraint on the contextual representation used in residual coupling.
During training, we jointly optimize a final emotion classification loss, a cross-modal contrastive learning loss, and an auxiliary classification loss on the contextual representation.

The final emotion classification loss is defined by cross-entropy:
\begin{equation}
\mathcal{L}_{\mathrm{cls}}
=
-\frac{1}{B}
\sum_{i=1}^{B}
\sum_{c=1}^{C}
y_{i,c}
\log \hat{p}_{i,c},
\label{eq:cls-loss}
\end{equation}
where $B$ denotes the batch size, $C$ is the number of emotion categories, $y_{i,c}$ is the ground-truth label indicator for the $i$-th utterance and class $c$, and $\hat{p}_{i,c}$ is the predicted probability of class $c$ in the distribution $\hat{\mathbf{p}}_i$ of the $i$-th target utterance.

To align the affective semantics of the textual, acoustic, and contextual representations, we add cross-modal contrastive learning: the textual feature $t$, the utterance-level acoustic representation $\bar{A}$, and the contextual representation $v$ are mapped into a contrastive space by independent projection heads.
Directional InfoNCE losses are then computed between representations of different types from the same utterance, which gives $\mathcal{L}_{\mathrm{CL}}$.
To preserve the emotion-discriminative information of the projected contextual representation, we add an auxiliary emotion classifier on $v$ with loss $\mathcal{L}_v$, which is used only during training and removed at inference.
The full definitions of both losses are given in Appendix~\ref{app:objective}.
The overall training objective is
\begin{equation}
\mathcal{L}_{\mathrm{total}}
=
\mathcal{L}_{\mathrm{cls}}
+
\lambda_{\mathrm{CL}}
\mathcal{L}_{\mathrm{CL}}
+
\lambda_{\mathrm{aux}}
\mathcal{L}_v,
\label{eq:total-loss}
\end{equation}
where $\lambda_{\mathrm{CL}}$ and $\lambda_{\mathrm{aux}}$ control the weights of the cross-modal contrastive learning loss and the contextual representation auxiliary loss, respectively, with their values given in Table~\ref{tab:fusion-settings} in Appendix~\ref{app:impl}.


\subsection{Training Procedure}
\label{sec:training-procedure}

We separately extract the contextual representation from the adapted VLM and the textual and acoustic features from RoBERTa-large and Emotion2vec-large, and cache them for subsequent training.
During training, the upstream feature extractors remain frozen, while the feature mappings, acoustic interaction, residual coupling, contrastive projection heads, and classifiers are optimized.
The complete training procedure is given in Algorithm~\ref{alg:training} in Appendix~\ref{app:training}.
\section{Experiments}
\label{sec:experiments}

\subsection{Experimental Setup}
\label{sec:experimental-setup}

\subsubsection{Datasets and Evaluation}

We evaluate \method{} on two widely used multimodal ERC benchmarks, IEMOCAP~\cite{busso2008iemocap} and MELD~\cite{poria2019meld}.
For IEMOCAP, we use the six-class setting: Happy, Sad, Neutral, Angry, Excited, and Frustrated.
Following a session-based evaluation protocol, we split Sessions 1--4 into training and development sets and hold out the entire Session 5 as an independent test set.
For MELD, we use the seven-class setting and the official training, development, and test splits.
The split sizes are given in Table~\ref{tab:dataset-splits}, and the class-wise support in the test sets is given in Table~\ref{tab:class-support} (Appendix~\ref{app:datasets}).
Since IEMOCAP and MELD are class-imbalanced, overall accuracy alone may not reflect how well a model recognizes each emotion category.
We therefore report accuracy (Acc.), weighted F1 (W-F1), macro-averaged F1 (M-F1), and class-wise F1 together.
Among them, M-F1 gives all emotion classes equal weight and thus better reflects performance on less-represented classes. The metric definitions are given in Appendix~\ref{app:datasets}.

\subsubsection{Implementation Details}
\label{sec:implementation-details}

The training procedure consists of two stages: emotion-supervised VLM adaptation and subsequent local-cue coupling for multimodal emotion recognition.
The corresponding training hyperparameters for both stages are provided separately in Appendix~\ref{app:impl}.


\subsection{Main Results}
\label{sec:main-results}

We compare \method{} with two groups of existing methods on the IEMOCAP and MELD datasets: conventional multimodal ERC methods and LLM/MLLM-based ERC methods.
For conventional methods with publicly available source code, we reproduce their results using the same data splits and five random seeds as \method{}, reporting the mean $\pm$ standard deviation over five runs.
For the remaining methods, we use the results reported in their original publications.
The methods not reproduced in this study follow the same test-set protocols as \method{}.
On MELD, all methods use the official training, development, and test splits.
On IEMOCAP, the compared methods also use Session 5 as the held-out test set, with training and development data drawn from Sessions 1--4.

\begin{table*}[t]
\centering
\caption{
Comparison with Representative ERC Methods on IEMOCAP.
Best Results Are Highlighted in Bold.
}
\label{tab:iemocap-results}

\scriptsize
\setlength{\tabcolsep}{3pt}
\renewcommand{\arraystretch}{1.05}

\resizebox{\textwidth}{!}{
\begin{tabular}{lccccccccc}
\toprule

Method
& Happy
& Sad
& Neutral
& Angry
& Excited
& Frustrated
& Acc.
& W-F1
& M-F1 \\

\midrule

\multicolumn{10}{l}{
\textit{Conventional / Multimodal ERC Methods}
} \\

MM-DFN~\cite{hu2022mm} (ICASSP'22)
& 42.22 & 78.98 & 66.42
& 69.77 & 75.56 & 66.33
& -- & 68.18 & 66.55 \\

M2FNet~\cite{chudasama2022m2fnet} (CVPRW'22)
& 60.00 & 82.11 & 65.88
& 68.21 & 72.60 & 68.31
& -- & 69.86 & 69.52 \\

PCGNet~\cite{tu2024pcgnet} (SIGIR'24)$^\ddagger$
& $47.06\pm1.85$
& $84.16\pm0.45$
& $69.19\pm1.03$
& $66.32\pm0.93$
& $73.12\pm2.10$
& $67.83\pm1.20$
& $69.60\pm0.31$
& $69.59\pm0.24$
& $67.95\pm0.33$ \\

CMERC~\cite{tu2024cmerc} (ACM MM'24)$^\ddagger$
& $53.41\pm3.90$
& $81.20\pm3.65$
& $68.98\pm0.91$
& $65.97\pm1.93$
& $73.31\pm2.61$
& $65.80\pm2.37$
& $68.99\pm1.02$
& $69.18\pm0.92$
& $68.11\pm1.07$ \\

DER-GCN~\cite{ai2025dergcn} (TNNLS'25)
& 58.80 & 79.80 & 61.50
& 72.10 & 73.30 & 67.80
& -- & 68.80 & 68.88 \\

ECERC~\cite{zhang-tan-2025-ecerc} (ACL'25)$^\ddagger$
& $58.19\pm1.59$
& $78.17\pm1.52$
& $67.24\pm0.65$
& $64.20\pm0.62$
& $67.73\pm4.00$
& $66.11\pm1.44$
& $67.31\pm0.61$
& $67.60\pm0.67$
& $66.94\pm0.67$ \\

CSS~\cite{lyu2026css} (AAAI'26)$^\ddagger$
& $57.60\pm1.40$
& $81.77\pm0.34$
& $73.76\pm0.86$
& $67.30\pm1.58$
& $71.70\pm2.79$
& $69.05\pm0.63$
& $71.11\pm0.51$
& $71.38\pm0.58$
& $70.20\pm0.55$ \\

EMART~\cite{seong2026emart} (ACL'26)
& 60.42
& 79.59
& \textbf{83.07}
& 61.17
& 66.22
& 64.57
& 70.79
& 70.93
& 69.17 \\

\midrule

\multicolumn{10}{l}{
\textit{LLM / MLLM-based ERC Methods}
} \\

InstructERC~\cite{lei2023instructerc} (arXiv'23)
& -- & -- & -- & -- & -- & --
& -- & 71.39 & -- \\

DialogueLLM~\cite{zhang2025dialoguellm} (Neural Networks'25)
& -- & -- & -- & -- & -- & --
& 70.62 & 69.93 & -- \\

SpeechCueLLM~\cite{wu2025beyond} (NAACL Findings'25)
& -- & -- & -- & -- & -- & --
& -- & 72.60 & -- \\

GRPO-LLM~\cite{chen-etal-2025-grpo} (EMNLP Findings'25)
& -- & -- & -- & -- & -- & --
& 72.95 & 73.04 & -- \\

DialogueMMT~\cite{he2025dialoguemmt} (COLING'25)
& 64.10 & 83.51 & 71.37
& 68.70 & 72.20 & 69.99
& 72.58 & 72.71 & 71.65 \\

BeMERC~\cite{fu2025bemerc} (arXiv'25)
& 59.42 & 84.54 & 75.20
& 69.39 & 79.87 & \textbf{72.70}
& 74.98 & 74.88 & 73.52 \\

Causal-ERC~\cite{jing2026causal} (AAAI'26)
& 62.95 & 80.61 & 72.80
& 67.76 & 74.14 & 68.83
& 71.66 & 71.89 & 71.18 \\

PRC-Emo~\cite{li2026prcemo} (AAAI'26)
& -- & -- & -- & -- & -- & --
& 71.03 & 71.95 & -- \\

\midrule

\textbf{\method{}}
& $\mathbf{68.17\pm1.05}$
& $\mathbf{86.61\pm0.90}$
& $78.82\pm0.53$
& $\mathbf{72.15\pm0.59}$
& $\mathbf{81.38\pm0.44}$
& $71.31\pm0.40$
& $\mathbf{76.99\pm0.23}$
& $\mathbf{77.07\pm0.22}$
& $\mathbf{76.41\pm0.26}$ \\

\bottomrule
\end{tabular}
}

\par\smallskip
\begin{minipage}{0.98\textwidth}
\scriptsize
$^\ddagger$ denotes results reproduced in this study.
A dash indicates that the corresponding result is not reported or unavailable.
\end{minipage}

\end{table*}

As shown in Table~\ref{tab:iemocap-results}, \method{} reaches 76.99\% accuracy, 77.07\% W-F1, and 76.41\% M-F1 on IEMOCAP, the best result on all three overall metrics among the compared methods that report them.
\begin{table*}[t]
\centering
\caption{
Comparison with Representative ERC Methods on MELD.
Best Results Are Highlighted in Bold.
}
\label{tab:meld-results}

\scriptsize
\setlength{\tabcolsep}{3pt}
\renewcommand{\arraystretch}{1.05}

\resizebox{\textwidth}{!}{
\begin{tabular}{lcccccccccc}
\toprule

Method
& Neutral
& Surprise
& Fear
& Sadness
& Joy
& Disgust
& Anger
& Acc.
& W-F1
& M-F1 \\

\midrule

\multicolumn{11}{l}{
\textit{Conventional / Multimodal ERC Methods}
} \\

MM-DFN~\cite{hu2022mm} (ICASSP'22)
& 77.38 & 57.19 & 14.29 & 41.36
& 63.25 & 20.41 & 51.44
& -- & 64.04 & 46.47 \\

M2FNet~\cite{chudasama2022m2fnet} (CVPRW'22)
& 80.06 & 58.66 & 3.45 & 47.03
& 65.50 & 25.24 & 55.25
& -- & 66.71 & 47.88 \\

PCGNet~\cite{tu2024pcgnet} (SIGIR'24)$^\ddagger$
& $79.37\pm0.45$
& $58.80\pm1.55$
& $16.89\pm3.78$
& $41.03\pm1.48$
& $64.47\pm0.55$
& $19.56\pm6.83$
& $53.39\pm1.43$
& $66.56\pm0.67$
& $65.62\pm0.47$
& $47.64\pm1.25$ \\

CMERC~\cite{tu2024cmerc} (ACM MM'24)$^\ddagger$
& $79.30\pm0.20$
& $58.68\pm0.94$
& $19.87\pm7.57$
& $40.54\pm1.10$
& $63.69\pm0.36$
& $21.93\pm2.04$
& $53.24\pm1.06$
& $66.18\pm0.41$
& $65.51\pm0.12$
& $48.18\pm1.04$ \\

DER-GCN~\cite{ai2025dergcn} (TNNLS'25)
& 80.60 & 51.00 & 10.40 & 41.50
& 64.30 & 10.30 & 57.40
& -- & 65.50 & 45.07 \\

ECERC~\cite{zhang-tan-2025-ecerc} (ACL'25)$^\ddagger$
& $79.33\pm0.46$
& $58.23\pm0.85$
& $21.33\pm3.27$
& $40.27\pm1.20$
& $63.64\pm1.11$
& $31.00\pm2.55$
& $52.44\pm1.85$
& $66.48\pm0.55$
& $65.60\pm0.53$
& $49.46\pm0.64$ \\

CSS~\cite{lyu2026css} (AAAI'26)$^\ddagger$
& $79.45\pm0.14$
& $57.07\pm0.60$
& $19.66\pm1.91$
& $40.67\pm0.52$
& $63.04\pm0.31$
& $28.27\pm1.98$
& $53.25\pm0.38$
& $66.47\pm0.22$
& $65.48\pm0.25$
& $48.77\pm0.68$ \\

EMART~\cite{seong2026emart} (ACL'26)
& \textbf{87.74}
& 46.62
& 20.59
& 27.88
& 60.45
& 4.41
& 52.12
& 66.70
& 64.71
& 42.83 \\

\midrule

\multicolumn{11}{l}{
\textit{LLM / MLLM-based ERC Methods}
} \\

InstructERC~\cite{lei2023instructerc} (arXiv'23)
& -- & -- & -- & -- & -- & -- & --
& -- & 69.15 & -- \\

DialogueLLM~\cite{zhang2025dialoguellm} (Neural Networks'25)
& -- & -- & -- & -- & -- & -- & --
& \textbf{71.96}
& \textbf{71.90}
& -- \\

SpeechCueLLM~\cite{wu2025beyond} (NAACL Findings'25)
& -- & -- & -- & -- & -- & -- & --
& -- & 67.60 & -- \\

GRPO-LLM~\cite{chen-etal-2025-grpo} (EMNLP Findings'25)
& -- & -- & -- & -- & -- & -- & --
& 71.00 & 70.09 & -- \\

DialogueMMT~\cite{he2025dialoguemmt} (COLING'25)
& 82.55
& 63.59
& 33.68
& 49.02
& 66.75
& 38.76
& \textbf{62.39}
& 71.19
& 70.66
& 56.68 \\

BeMERC~\cite{fu2025bemerc} (arXiv'25)
& 82.33
& \textbf{64.49}
& 32.10
& 44.03
& 66.83
& 30.23
& 60.66
& 71.18
& 69.78
& 54.38 \\

Causal-ERC~\cite{jing2026causal} (AAAI'26)
& 81.84
& 62.96
& 31.75
& 47.67
& \textbf{67.69}
& \textbf{45.53}
& 61.03
& 70.50
& 70.25
& 56.92 \\

PRC-Emo~\cite{li2026prcemo} (AAAI'26)
& -- & -- & -- & -- & -- & -- & --
& 71.50 & 70.44 & -- \\

\midrule

\textbf{\method{}}
& $81.96\pm0.36$
& $58.91\pm1.15$
& $\mathbf{39.61\pm0.89}$
& $\mathbf{50.14\pm1.35}$
& $66.75\pm0.44$
& $43.41\pm3.59$
& $61.43\pm0.83$
& $70.43\pm0.45$
& $70.07\pm0.37$
& $\mathbf{57.46\pm0.52}$ \\

\bottomrule
\end{tabular}
}

\par\smallskip
\begin{minipage}{0.98\textwidth}
\scriptsize
$^\ddagger$ denotes results reproduced in this study.
A dash indicates that the corresponding result is not reported or unavailable.
\end{minipage}

\end{table*}
On MELD, as shown in Table~\ref{tab:meld-results}, \method{} reaches 70.43\% accuracy, 70.07\% W-F1, and 57.46\% M-F1.
Among the compared methods with complete class-wise results, \method{} has the highest M-F1.
In particular, it obtains F1 scores of 39.61\% and 50.14\% on Fear and Sadness, which have only 50 and 208 test instances, respectively.
Its Acc. and W-F1 remain competitive with several LLM/MLLM-based methods, although they are not the highest.
Since the Neutral class makes up 48.1\% of the MELD test set, both metrics depend heavily on this majority class.


\subsection{Context and Local Cue Ablations}
\label{sec:ablation}

We design two groups of ablation experiments.
The first group examines how different VLM input sources affect the discriminative capability of the contextual representation.
The second group investigates whether local textual and acoustic cues provide additional discriminative information beyond the contextual representation.
The results are reported in Table~\ref{tab:evidence-ablation}.

\subsubsection{Contextual Input Ablation}

We examine the contribution of different contextual inputs to downstream emotion discrimination by removing dialogue history, facial frames, or audio descriptions from the adapted VLM.
The LoRA parameters remain fixed without re-adaptation.
As shown in Table~\ref{tab:evidence-ablation}, removing any input source reduces recognition performance.
Dialogue history has the largest impact.
Its removal reduces M-F1 by 5.94 points on MELD, compared with 2.07 points on IEMOCAP.
Facial frames and audio descriptions also contribute to recognition.
These results highlight the role of multiple contextual sources in supporting emotion discrimination through the adapted contextual representation.

\subsubsection{Local Cue Ablation}

We next examine whether local textual and acoustic cues extracted from the target utterance add discriminative value beyond the task-adapted contextual representation.
We remove one local branch at a time while retaining the contextual representation and the other cue.
As shown in Table~\ref{tab:evidence-ablation}, removing either local branch reduces M-F1 on both datasets.
However, the relative contributions differ.
Acoustic cues have a larger marginal contribution on IEMOCAP, whereas textual cues contribute slightly more on MELD.
These results show that incorporating additional local textual and acoustic cues alongside the VLM-derived contextual representation improves emotion recognition.

\subsubsection{Dataset-Level Characteristics of Local Modalities}

The preceding ablation results show different contributions of local modalities across the two datasets.
To investigate this difference, we evaluate unimodal classifiers using RoBERTa textual and Emotion2vec acoustic representations, following the same five-seed protocol.

\begin{table}[!t]
\centering

\caption{
    Unimodal classification on IEMOCAP and MELD.
}
\label{tab:unimodal-local-cues}

\footnotesize
\setlength{\tabcolsep}{1pt}
\renewcommand{\arraystretch}{1.12}

\begin{tabular*}{\columnwidth}{@{\extracolsep{\fill}}llccc@{}}
\toprule

\textbf{Dataset}
& \textbf{Modality}
& \textbf{Acc.}
& \textbf{W-F1}
& \textbf{M-F1} \\

\midrule

\multirow{2}{*}{IEMOCAP}
& RoBERTa-only
& $67.21\pm0.47$
& $67.10\pm0.38$
& $65.88\pm0.29$ \\

& Emotion2vec-only
& $67.30\pm0.47$
& $65.32\pm0.19$
& $62.63\pm0.18$ \\

\midrule

\multirow{2}{*}{MELD}
& RoBERTa-only
& $65.41\pm0.34$
& $65.29\pm0.25$
& $48.96\pm0.53$ \\

& Emotion2vec-only
& $50.06\pm0.28$
& $47.20\pm0.39$
& $27.88\pm0.61$ \\

\bottomrule
\end{tabular*}

\end{table}

As shown in Table~\ref{tab:unimodal-local-cues}, textual and acoustic features achieve M-F1 scores of 65.88\% and 62.63\% on IEMOCAP, respectively, compared with 48.96\% and 27.88\% on MELD.
The larger gap on MELD indicates dataset-dependent differences in the discriminative capabilities of local modalities.
However, unimodal discriminative capability does not directly determine its contribution to fusion.
Removing acoustic cues causes a larger performance drop on IEMOCAP (3.48 vs. 1.78 percentage points), while removing textual cues has a slightly greater impact on MELD (1.21 vs. 0.95 percentage points).
This suggests that fusion contributions depend not only on standalone discriminative capability but also on the complementary information provided beyond the contextual representation.

\begin{table*}[!t]
\centering

\caption{
Contextual Input and Local Cue Ablations on IEMOCAP and MELD. 
}
\label{tab:evidence-ablation}

\small
\setlength{\tabcolsep}{3pt}
\renewcommand{\arraystretch}{1.06}

\begin{tabular*}{\textwidth}{@{\extracolsep{\fill}}lcccccc@{}}
\toprule

\multirow{2}{*}{\textbf{Variant}}
& \multicolumn{3}{c}{\textbf{IEMOCAP}}
& \multicolumn{3}{c}{\textbf{MELD}} \\

\cmidrule(lr){2-4}
\cmidrule(lr){5-7}

& \textbf{Acc.}
& \textbf{W-F1}
& \textbf{M-F1}
& \textbf{Acc.}
& \textbf{W-F1}
& \textbf{M-F1} \\

\midrule

\multicolumn{7}{l}{\textit{Contextual Input Ablation}} \\

w/o VLM Dialogue History
& $75.61\pm0.58$
& $75.52\pm0.62$
& $74.34\pm0.70$
& $67.39\pm0.47$
& $67.01\pm0.57$
& $51.52\pm1.44$ \\

w/o VLM Audio Description
& $76.80\pm0.54$
& $76.78\pm0.54$
& $75.69\pm0.83$
& $70.19\pm0.50$
& $69.93\pm0.37$
& $56.63\pm0.46$ \\

w/o VLM Facial Frames
& $76.06\pm0.46$
& $76.08\pm0.51$
& $75.15\pm0.80$
& $69.88\pm0.72$
& $69.50\pm0.66$
& $55.65\pm1.25$ \\

\midrule

\multicolumn{7}{l}{\textit{Local Cue Ablation}} \\

w/o Local Acoustic Cues
& $73.49\pm0.22$
& $73.55\pm0.28$
& $72.93\pm0.50$
& $69.58\pm0.14$
& $69.31\pm0.10$
& $56.51\pm0.26$ \\

w/o Local Textual Cues
& $75.02\pm0.25$
& $75.17\pm0.28$
& $74.63\pm0.31$
& $69.37\pm0.13$
& $69.16\pm0.08$
& $56.25\pm0.18$ \\

\midrule

\textbf{Full \method{}}
& $\mathbf{76.99\pm0.23}$
& $\mathbf{77.07\pm0.22}$
& $\mathbf{76.41\pm0.26}$
& $\mathbf{70.43\pm0.45}$
& $\mathbf{70.07\pm0.37}$
& $\mathbf{57.46\pm0.52}$ \\

\bottomrule
\end{tabular*}

\end{table*}

\subsubsection{Class-Wise Complementarity of Local Cues}

\begin{figure*}[!t]
    \centering
    \begin{minipage}[t]{0.49\textwidth}
        \centering
        \vspace{0pt}
        \includegraphics[width=\linewidth]{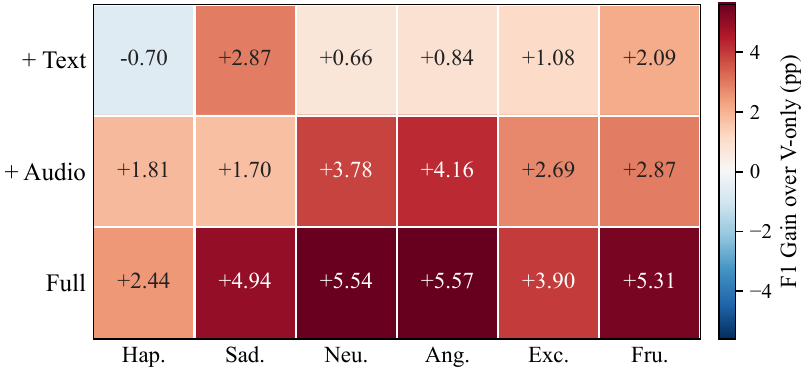}
        \par\vspace{2pt}
        {\small (a) IEMOCAP}
    \end{minipage}\hfill%
    \begin{minipage}[t]{0.49\textwidth}
        \centering
        \vspace{0pt}
        \includegraphics[width=\linewidth]{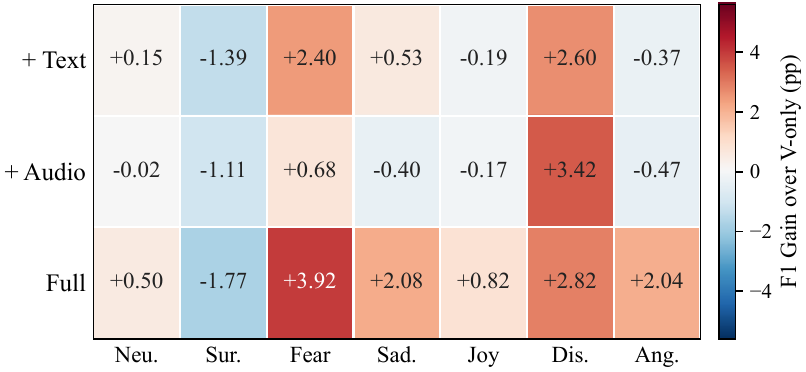}
        \par\vspace{2pt}
        {\small (b) MELD}
    \end{minipage}
    \caption{
        Class-wise F1 changes relative to V-only after incorporating local textual and acoustic cues.
    }
    \label{fig:classwise-complementarity}
\end{figure*}

To further examine the complementarity of local cues, we analyze class-wise F1 changes on both IEMOCAP and MELD relative to V-only.
Textual and acoustic cues are added separately or jointly.
As shown in Fig.~\ref{fig:classwise-complementarity}, Full \method{} improves F1 across all six emotion categories on IEMOCAP.
Neutral, Angry, and Frustrated gain 5.54, 5.57, and 5.31 points, respectively.
In contrast, the gains on MELD are more class-dependent.
Fear and Disgust improve by 3.92 and 2.82 points, while Surprise decreases by 1.77 points.

These results indicate that local cues complement the task-adapted contextual representation.
However, their contributions vary across datasets and emotion categories.
The overall M-F1 comparison is visualized in Appendix~\ref{app:complementarity}.

\subsection{VLM Adaptation Analysis}
\label{sec:representation-analysis}

We next examine how emotion-supervised adaptation affects the discriminative capability of continuous VLM representations.
We compare the frozen pretrained VLM with its ERC-supervised LoRA-adapted counterpart.
Apart from LoRA adaptation, both settings use identical input configurations, representation extraction procedures, and downstream classifier settings.
We train downstream classifiers on their respective continuous representations for emotion recognition at the target-utterance level.

\begin{table}[t]
\centering

\caption{
Effect of Emotion-Supervised VLM Adaptation on Classification Using the Continuous Internal Representation.
}
\label{tab:vlm-adaptation-effect}

\footnotesize
\setlength{\tabcolsep}{3pt}
\renewcommand{\arraystretch}{1.08}

\resizebox{\columnwidth}{!}{
\begin{tabular}{llccc}
\toprule

\textbf{Dataset}
& \textbf{VLM}
& \textbf{Acc.}
& \textbf{W-F1}
& \textbf{M-F1} \\

\midrule

\multirow{2}{*}{IEMOCAP}
& Frozen
& $64.12\pm0.95$
& $64.27\pm1.03$
& $62.51\pm1.24$ \\

& LoRA
& $\mathbf{72.08\pm0.19}$
& $\mathbf{72.25\pm0.17}$
& $\mathbf{71.79\pm0.16}$ \\

\midrule

\multirow{2}{*}{MELD}
& Frozen
& $59.99\pm0.30$
& $58.55\pm0.33$
& $42.51\pm0.23$ \\

& LoRA
& $\mathbf{69.75\pm0.28}$
& $\mathbf{69.31\pm0.26}$
& $\mathbf{55.97\pm0.48}$ \\

\bottomrule
\end{tabular}
}

\end{table}

As shown in Table~\ref{tab:vlm-adaptation-effect}, emotion-supervised adaptation substantially improves downstream classification performance.
M-F1 increases by 9.28 and 13.46 percentage points on IEMOCAP and MELD, respectively.
Accuracy and W-F1 also improve, indicating that ERC-supervised adaptation makes the internal representations more suitable for emotion discrimination.
We further compare direct label generation with representation-based classification using the same adapted VLM.
Both methods use identical trained VLM parameters to avoid differences caused by model adaptation.
We select one of the five random-seed runs for this comparison.

\begin{figure*}[!t]
    \centering
    \begin{minipage}[t]{0.48\textwidth}
        \centering
        \vspace{0pt}
        \includegraphics[width=\linewidth]{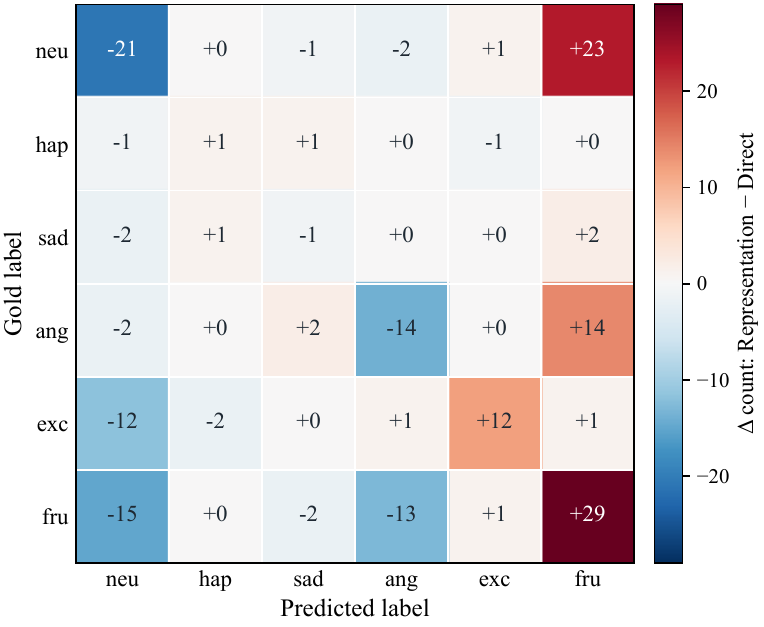}
        \par\vspace{2pt}
        {\small (a) IEMOCAP}
    \end{minipage}\hfill%
    \begin{minipage}[t]{0.48\textwidth}
        \centering
        \vspace{0pt}
        \includegraphics[width=\linewidth]{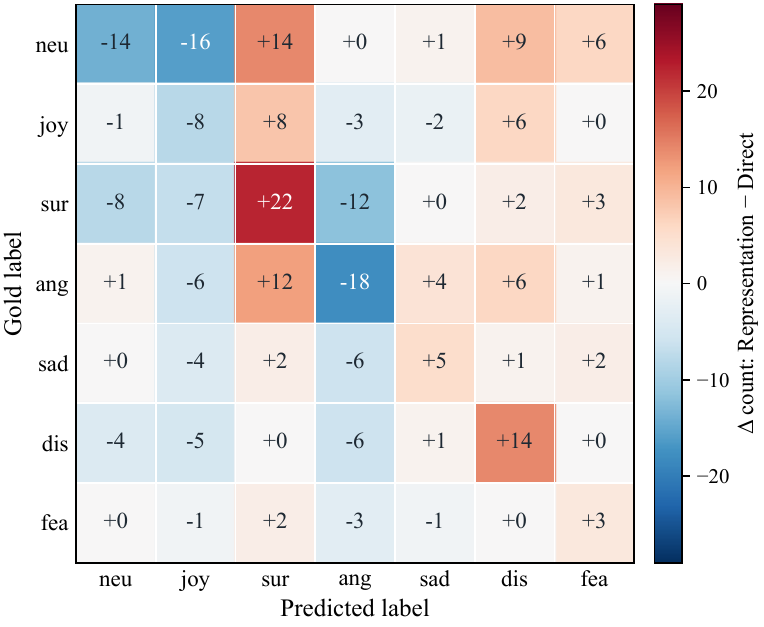}
        \par\vspace{2pt}
        {\small (b) MELD}
    \end{minipage}
    \caption{
        Prediction differences between representation-based classification and direct label generation.
    }
    \label{fig:representation-comparison}
    \label{fig:representation-confusion-iemocap}
    \label{fig:representation-confusion-meld}
\end{figure*}

As shown in Fig.~\ref{fig:representation-comparison}, the two prediction methods exhibit class-dependent differences on both datasets.
Each matrix shows the difference in prediction counts between representation-based classification and direct label generation.
Positive and negative values indicate increases and decreases in the corresponding counts, respectively.
These differences show that representation-based classification does not simply reproduce the discrete emotion labels generated by the VLM.


\subsection{Fusion Mechanism Analysis}
\label{sec:fusion-comparison}

The local cue ablations in Section~\ref{sec:ablation} confirm the benefits of incorporating additional local textual and acoustic cues.
However, they do not isolate the contributions of local features and the proposed coupling mechanism.
We therefore examine residual construction, sample-adaptive gating, and alternative fusion strategies.
All variants use the same feature extractors, input representations, and training protocol.
The results are reported in Table~\ref{tab:fusion-comparison}.
Additional objective ablations are provided in Appendix~\ref{app:objective-ablation}.

\begin{table}[!t]
\centering

\caption{
    Fusion-Mechanism Comparisons.
}
\label{tab:fusion-comparison}

\footnotesize
\setlength{\tabcolsep}{1pt}
\renewcommand{\arraystretch}{1.12}

\begin{tabular*}{\columnwidth}{@{\extracolsep{\fill}}llccc@{}}
\toprule

\textbf{Dataset}
& \textbf{Method}
& \textbf{Acc.}
& \textbf{W-F1}
& \textbf{M-F1} \\

\midrule

\multirow{5}{*}{IEMOCAP}
& Non-Residual
& $75.57\pm0.41$
& $75.60\pm0.42$
& $74.70\pm0.49$ \\

& Global Gating
& $75.60\pm0.13$
& $75.59\pm0.10$
& $74.57\pm0.20$ \\

& Concat-MLP
& $76.15\pm0.11$
& $76.15\pm0.18$
& $75.31\pm0.36$ \\

& Weighted Fusion
& $75.93\pm0.35$
& $75.95\pm0.35$
& $75.06\pm0.51$ \\

& \textbf{Full \method{}}
& $\mathbf{76.99\pm0.23}$
& $\mathbf{77.07\pm0.22}$
& $\mathbf{76.41\pm0.26}$ \\

\midrule

\multirow{5}{*}{MELD}
& Non-Residual
& $69.42\pm0.28$
& $69.10\pm0.22$
& $56.25\pm0.48$ \\

& Global Gating
& $68.87\pm0.21$
& $68.72\pm0.18$
& $54.56\pm0.50$ \\

& Concat-MLP
& $69.04\pm0.50$
& $68.81\pm0.38$
& $55.93\pm0.38$ \\

& Weighted Fusion
& $69.51\pm0.38$
& $69.32\pm0.35$
& $56.50\pm0.68$ \\

& \textbf{Full \method{}}
& $\mathbf{70.43\pm0.45}$
& $\mathbf{70.07\pm0.37}$
& $\mathbf{57.46\pm0.52}$ \\

\bottomrule
\end{tabular*}

\end{table}

\subsubsection{Residual Construction}

We first examine the effect of residual construction on the use of local cues.
Non-Residual retains the same local features and sample-adaptive gating.
It removes the difference terms relative to the contextual representation.
It directly gates and fuses the textual and context-guided acoustic representations.
Removing residual construction reduces M-F1 by 1.71 and 1.21 points on IEMOCAP and MELD, respectively, relative to the full model.
This indicates that residual construction helps exploit local discriminative cues more effectively.

\subsubsection{Global versus Sample-Adaptive Gating}

We next examine whether local cue contributions should vary across utterances.
Global Gating retains the residual structure but uses globally learned coefficients shared across utterances.
Compared with Global Gating, sample-adaptive gating improves M-F1 by 1.84 and 2.90 points on IEMOCAP and MELD, respectively.
These results support adjusting local cue contributions dynamically for each utterance rather than using shared weights.

\subsubsection{Comparison with Common Fusion Strategies}

We further compare our coupling design with two common fusion strategies using the same local features.
Concat-MLP concatenates the contextual, textual, and context-guided acoustic representations before applying an MLP.
Weighted Fusion directly combines the three representations using sample-dependent weights.
Full \method{} outperforms both fusion strategies on the two datasets.
It improves M-F1 over Concat-MLP by 1.10 and 1.53 points on IEMOCAP and MELD, respectively.
The corresponding M-F1 gains over Weighted Fusion are 1.35 and 0.96 points, respectively.
\section{Conclusion}
\label{sec:conclusion}

Accurately recognizing emotions in video calls and livestreamed discussions helps understand speakers' emotional expressions. To address two challenges in multimodal Emotion Recognition in Conversation (ERC), we propose CLEAR. To address the underutilization of internal affective information caused by the reliance of existing MLLM-based methods on generated discrete emotion labels, we enhance the emotion-discriminative capability of VLM representations through ERC-supervised LoRA adaptation, extract its continuous contextual representation, and train an additional classifier for emotion discrimination. To address the potential lack of salience of local discriminative cues from the target utterance in rich contextual modeling, we independently extract its textual and acoustic features, obtain local acoustic evidence through context-guided acoustic attention, and employ sample-adaptive residual coupling to dynamically integrate local evidence with the contextual representation, strengthening the discriminative role of local cues while preserving contextual information.
Experiments on IEMOCAP and MELD demonstrate the effectiveness of CLEAR. Further analyses support the two design choices: emotion-supervised VLM adaptation improves the Macro-F1 of representation-based classification by 9.28 and 13.46 percentage points on IEMOCAP and MELD, respectively. Incorporating utterance-specific textual and acoustic evidence into the adapted contextual representation brings additional gains of 4.62 and 1.49 points. 

These results provide methodological support for fine-grained emotion recognition in online conversational video content. As the current approach uses offline feature extraction and a bounded dialogue-history window, its efficiency and robustness in longer, dynamically evolving conversations remain to be examined. In future work, we will explore emotion dynamics modeling and more efficient feature processing for long-form conversational videos to further support fine-grained online video content analysis.

\bibliographystyle{ACM-Reference-Format}
\bibliography{refer}

@inproceedings{poria2019meld,
  title={Meld: A multimodal multi-party dataset for emotion recognition in conversations},
  author={Poria, Soujanya and Hazarika, Devamanyu and Majumder, Navonil and Naik, Gautam and Cambria, Erik and Mihalcea, Rada},
  booktitle={Proceedings of the 57th annual meeting of the association for computational linguistics},
  pages={527--536},
  year={2019}
}

@inproceedings{tu2024pcgnet,
  title={A Persona-Infused Cross-Task Graph Network for Multimodal Emotion Recognition with Emotion Shift Detection in Conversations},
  author={Tu, Geng and Xiong, Feng and Liang, Bin and Xu, Ruifeng},
  booktitle={Proceedings of the 47th International ACM SIGIR Conference on Research and Development in Information Retrieval},
  pages={2266--2270},
  year={2024},
  publisher={ACM},
  doi={10.1145/3626772.3657944}
}

@inproceedings{tu2024cmerc,
  title={Multimodal Emotion Recognition Calibration in Conversations},
  author={Tu, Geng and Xiong, Feng and Liang, Bin and Wang, Hui and Zeng, Xi and Xu, Ruifeng},
  booktitle={Proceedings of the 32nd ACM International Conference on Multimedia},
  year={2024},
  publisher={ACM},
  doi={10.1145/3664647.3681515}
}

@article{busso2008iemocap,
  title={IEMOCAP: Interactive emotional dyadic motion capture database},
  author={Busso, Carlos and Bulut, Murtaza and Lee, Chi-Chun and Kazemzadeh, Abe and Mower, Emily and Kim, Samuel and Chang, Jeannette N and Lee, Sungbok and Narayanan, Shrikanth S},
  journal={Language resources and evaluation},
  volume={42},
  number={4},
  pages={335--359},
  year={2008},
  publisher={Springer}
}

@inproceedings{ghosal2019dialoguegcn,
  title={Dialoguegcn: A graph convolutional neural network for emotion recognition in conversation},
  author={Ghosal, Deepanway and Majumder, Navonil and Poria, Soujanya and Chhaya, Niyati and Gelbukh, Alexander},
  booktitle={Proceedings of the 2019 conference on empirical methods in natural language processing and the 9th International Joint Conference on Natural Language Processing (EMNLP-IJCNLP)},
  pages={154--164},
  year={2019}
}

@inproceedings{hu2022mm,
  title={MM-DFN: Multimodal dynamic fusion network for emotion recognition in conversations},
  author={Hu, Dou and Hou, Xiaolong and Wei, Lingwei and Jiang, Lianxin and Mo, Yang},
  booktitle={ICASSP 2022-2022 IEEE International Conference on Acoustics, Speech and Signal Processing (ICASSP)},
  pages={7037--7041},
  year={2022},
  organization={IEEE}
}

@inproceedings{chudasama2022m2fnet,
  title={{M2FNet}: Multi-Modal Fusion Network for Emotion Recognition in Conversation},
  author={Chudasama, Vishal and Kar, Purbayan and Gudmalwar, Ashish
          and Shah, Nirmesh and Wasnik, Pankaj and Onoe, Naoyuki},
  booktitle={Proceedings of the IEEE/CVF Conference on Computer Vision
             and Pattern Recognition (CVPR) Workshops},
  pages={4652--4661},
  year={2022}
}

@inproceedings{shi2023multiemo,
  title={MultiEMO: An attention-based correlation-aware multimodal fusion framework for emotion recognition in conversations},
  author={Shi, Tao and Huang, Shao-Lun},
  booktitle={Proceedings of the 61st Annual Meeting of the Association for Computational Linguistics (Volume 1: Long Papers)},
  pages={14752--14766},
  year={2023}
}

@article{lei2023instructerc,
  title={Instructerc: Reforming emotion recognition in conversation with multi-task retrieval-augmented large language models},
  author={Lei, Shanglin and Dong, Guanting and Wang, Xiaoping and Wang, Keheng and Qiao, Runqi and Wang, Sirui},
  journal={arXiv preprint arXiv:2309.11911},
  year={2023}
}

@article{zhang2025dialoguellm,
  title={Dialoguellm: Context and emotion knowledge-tuned large language models for emotion recognition in conversations},
  author={Zhang, Yazhou and Wang, Mengyao and Wu, Youxi and Tiwari, Prayag and Li, Qiuchi and Wang, Benyou and Qin, Jing},
  journal={Neural Networks},
  pages={107901},
  year={2025},
  publisher={Elsevier}
}

@inproceedings{he2025dialoguemmt,
  title={Dialoguemmt: Dialogue scenes understanding enhanced multi-modal multi-task tuning for emotion recognition in conversations},
  author={He, Chenyuan and Zhu, Senbin and Liu, Hongde and Gao, Fei and Jia, Yuxiang and Zan, Hongying and Peng, Min},
  booktitle={Proceedings of the 31st International Conference on Computational Linguistics},
  pages={2497--2512},
  year={2025}
}

@inproceedings{ma2024emotion2vec,
  title={emotion2vec: Self-supervised pre-training for speech emotion representation},
  author={Ma, Ziyang and Zheng, Zhisheng and Ye, Jiaxin and Li, Jinchao and Gao, Zhifu and Zhang, Shiliang and Chen, Xie},
  booktitle={Findings of the Association for Computational Linguistics: ACL 2024},
  pages={15747--15760},
  year={2024}
}

@inproceedings{zhang-tan-2025-ecerc,
  title={ECERC: evidence-cause attention network for multi-modal emotion recognition in conversation},
  author={Zhang, Tao and Tan, Zhenhua},
  booktitle={Proceedings of the 63rd Annual Meeting of the Association for Computational Linguistics (Volume 1: Long Papers)},
  pages={2064--2077},
  year={2025}
}

@inproceedings{majumder2019dialoguernn,
  title={{DialogueRNN}: An Attentive {RNN} for Emotion Detection in Conversations},
  author={Majumder, Navonil and Poria, Soujanya and Hazarika, Devamanyu and Mihalcea, Rada and Gelbukh, Alexander and Cambria, Erik},
  booktitle={Proceedings of the AAAI Conference on Artificial Intelligence},
  volume={33},
  number={01},
  pages={6818--6825},
  year={2019}
}

@inproceedings{aguilar2019multimodal,
  title={Multimodal and Multi-view Models for Emotion Recognition},
  author={Aguilar, Gustavo and Rozgic, Viktor and Wang, Weiran and Wang, Chao},
  booktitle={Proceedings of the 57th Annual Meeting of the Association for Computational Linguistics},
  pages={991--1002},
  year={2019},
  publisher={Association for Computational Linguistics},
  doi={10.18653/v1/P19-1095}
}

@inproceedings{shenoy2020multilogue,
  title={Multilogue-Net: A Context-Aware RNN for Multi-modal Emotion Detection and Sentiment Analysis in Conversation},
  author={Shenoy, Aman and Sardana, Ashish},
  booktitle={Second Grand-Challenge and Workshop on Multimodal Language},
  pages={19--28},
  year={2020},
  publisher={Association for Computational Linguistics},
  doi={10.18653/v1/2020.challengehml-1.3}
}

@inproceedings{lee2022compm,
  title={{CoMPM}: Context Modeling with Speaker's Pre-trained Memory Tracking for Emotion Recognition in Conversation},
  author={Lee, Joosung and Lee, Wooin},
  booktitle={Proceedings of the 2022 Conference of the North American Chapter of the Association for Computational Linguistics: Human Language Technologies},
  pages={5669--5679},
  year={2022}
}

@article{tu2024adagin,
  title={Adaptive Graph Learning for Multimodal Conversational Emotion Detection},
  author={Tu, Geng and Xie, Tian and Liang, Bin and Wang, Hongpeng and Xu, Ruifeng},
  journal={Proceedings of the AAAI Conference on Artificial Intelligence},
  volume={38},
  number={17},
  pages={19089--19097},
  year={2024}
}

@article{ai2025dergcn,
  title={{DER-GCN}: Dialog and Event Relation-Aware Graph Convolutional Neural Network for Multimodal Dialog Emotion Recognition},
  author={Ai, Wei and Shou, Yuntao and Meng, Tao and Li, Keqin},
  journal={IEEE Transactions on Neural Networks and Learning Systems},
  volume={36},
  number={3},
  pages={4908--4921},
  year={2025},
  doi={10.1109/TNNLS.2024.3367940}
}

@article{liu2024lost,
  title={Lost in the Middle: How Language Models Use Long Contexts},
  author={Liu, Nelson F. and Lin, Kevin and Hewitt, John and Paranjape, Ashwin and Bevilacqua, Michele and Petroni, Fabio and Liang, Percy},
  journal={Transactions of the Association for Computational Linguistics},
  volume={12},
  pages={157--173},
  year={2024},
  publisher={MIT Press},
  doi={10.1162/tacl_a_00638}
}

@inproceedings{yi2024haucl,
  title={Multimodal Fusion via Hypergraph Autoencoder and Contrastive Learning for Emotion Recognition in Conversation},
  author={Yi, Zijian and Zhao, Ziming and Shen, Zhishu and Zhang, Tiehua},
  booktitle={Proceedings of the 32nd ACM International Conference on Multimedia},
  pages={4341--4348},
  year={2024},
  doi={10.1145/3664647.3681633}
}

@inproceedings{dutta2025llm,
  title={Llm supervised pre-training for multimodal emotion recognition in conversations},
  author={Dutta, Soumya and Ganapathy, Sriram},
  booktitle={ICASSP 2025-2025 IEEE International Conference on Acoustics, Speech and Signal Processing (ICASSP)},
  pages={1--5},
  year={2025},
  organization={IEEE}
}

@inproceedings{wu2025beyond,
  title={Beyond silent letters: Amplifying llms in emotion recognition with vocal nuances},
  author={Wu, Zehui and Gong, Ziwei and Ai, Lin and Shi, Pengyuan and Donbekci, Kaan and Hirschberg, Julia},
  booktitle={Findings of the Association for Computational Linguistics: NAACL 2025},
  pages={2202--2218},
  year={2025}
}

@inproceedings{hu2022lora,
  title={LoRA: Low-Rank Adaptation of Large Language Models},
  author={Hu, Edward J and Shen, Yelong and Wallis, Phillip and
          Allen-Zhu, Zeyuan and Li, Yuanzhi and Wang, Shean and
          Wang, Lu and Chen, Weizhu},
  booktitle={International Conference on Learning Representations},
  year={2022}
}

@inproceedings{fu2025laerc,
  title={LaERC-S: Improving LLM-based emotion recognition in conversation with speaker characteristics},
  author={Fu, Yumeng and Wu, Junjie and Wang, Zhongjie and Zhang, Meishan and Shan, Lili and Wu, Yulin and Liu, Bingquan},
  booktitle={Proceedings of the 31st International Conference on Computational Linguistics},
  pages={6748--6761},
  year={2025}
}

@inproceedings{chen-etal-2025-grpo,
  title={{GRPO}-Guided Modality Selection Enhanced {L}o{RA}-Tuned {LLM}s for Multimodal Emotion Recognition},
  author={Chen, Yang and Yang, Shuwan and Xiang, Yan and Song, Ran and Huang, Yuxin and Yu, Zhengtao},
  booktitle={Findings of the Association for Computational Linguistics: EMNLP 2025},
  pages={19458--19471},
  year={2025},
  address={Suzhou, China},
  publisher={Association for Computational Linguistics},
  doi={10.18653/v1/2025.findings-emnlp.1059},
  url={https://aclanthology.org/2025.findings-emnlp.1059/}
}

@inproceedings{zhang2026mme,
  title={{MME-Emotion}: A Holistic Evaluation Benchmark for Emotional Intelligence in Multimodal Large Language Models},
  author={Zhang, Fan and Cheng, Zebang and Deng, Chong and Li, Haoxuan and Lian, Zheng and Chen, Qian and Liu, Huadai and Wang, Wen and Zhang, YiFan and Zhang, Renrui and Guo, Ziyu and Zhu, Zhihong and Wu, Hao and Wang, Haixin and Zheng, Yefeng and Peng, Xiaojiang and Wu, Xian and Wang, Kun and Li, Xiangang and Ye, Jieping and Heng, Pheng-Ann},
  booktitle={International Conference on Learning Representations},
  year={2026}
}

@article{fu2025bemerc,
  title   = {BeMERC: Behavior-Aware MLLM-based Framework for Multimodal Emotion Recognition in Conversation},
  author  = {Fu, Yumeng and Wu, Junjie and Wang, Zhongjie and Zhang, Meishan and Wu, Yulin and Liu, Bingquan},
  journal = {arXiv preprint arXiv:2503.23990},
  year    = {2025}
}

@article{lyu2026css,
  title={Cross-Space Synergy: A Unified Framework for Multimodal Emotion Recognition in Conversation},
  author={Lyu, Xiaosen and Xiong, Jiayu and Chen, Yuren and Wang, Wanlong and Dai, Xiaoqing and Wang, Jing},
  journal={Proceedings of the AAAI Conference on Artificial Intelligence},
  volume={40},
  number={29},
  pages={24226--24234},
  year={2026},
  doi={10.1609/aaai.v40i29.39602}
}

@article{jing2026causal,
  title={Causal-ERC: A Multimodal Framework with Causal Prompting for Emotion Recognition in Conversations with Large Language Models},
  author={Jing, Ran and Tu, Geng and Zhang, Yice and Xu, Ruifeng},
  journal={Proceedings of the AAAI Conference on Artificial Intelligence},
  volume={40},
  number={37},
  pages={31383--31391},
  year={2026},
  doi={10.1609/aaai.v40i37.40402}
}

@article{li2026prcemo,
  title={Do LLMs Feel? Teaching Emotion Recognition with Prompts, Retrieval, and Curriculum Learning},
  author={Li, Xinran and Liu, Yu and Qiao, Jiaqi and Xu, Xiujuan},
  journal={Proceedings of the AAAI Conference on Artificial Intelligence},
  volume={40},
  number={38},
  pages={31778--31786},
  year={2026},
  doi={10.1609/aaai.v40i38.40446}
}

@inproceedings{seong2026emart,
  title={Emotion-Wheel-Guided Audio-Referred Text Representation for Multimodal Emotion Recognition in Conversation},
  author={Seong, Eunseon and Lee, Harim and Kim, Dahye and Kim, Changhyun and Chae, Dong-Kyu},
  booktitle={Proceedings of the 64th Annual Meeting of the Association for Computational Linguistics (Volume 1: Long Papers)},
  pages={40392--40403},
  year={2026},
  publisher={Association for Computational Linguistics},
  doi={10.18653/v1/2026.acl-long.1875}
}

@inproceedings{sun2026emomm,
  title={EmoMM: Benchmarking and Steering MLLM for Multimodal Emotion Recognition under Conflict and Missingness},
  author={Sun, Yueru and Zhang, Yimeng and Gu, Haoyu and Chen, Nuo and She, Dong and Yao, Xianrong and Gao, Yang and Jin, Zhanpeng},
  booktitle={Findings of the Association for Computational Linguistics: ACL 2026},
  pages={20351--20371},
  year={2026}
}

@article{bai2025qwen3vl,
  title={Qwen3-VL Technical Report},
  author={Bai, Shuai and Cai, Yuxuan and Chen, Ruizhe and Chen, Keqin and others},
  journal={arXiv preprint arXiv:2511.21631},
  year={2025}
}

@article{liu2019roberta,
  title={RoBERTa: A Robustly Optimized BERT Pretraining Approach},
  author={Liu, Yinhan and Ott, Myle and Goyal, Naman and Du, Jingfei and Joshi, Mandar and Chen, Danqi and Levy, Omer and Lewis, Mike and Zettlemoyer, Luke and Stoyanov, Veselin},
  journal={arXiv preprint arXiv:1907.11692},
  year={2019}
}

@inproceedings{nguyen2026cmtd,
  title={CMTD: Cognitive Modeling with Traits and Distortions for Multimodal Emotion Recognition in Conversations},
  author={Nguyen, Minh-Tien and Le, Huu-Loi and Phan, Manh-Cuong and Hotta, Hajime},
  booktitle={Findings of the Association for Computational Linguistics: ACL 2026},
  pages={839--854},
  year={2026}
}

@article{chu2024qwen2audio,
  title={{Qwen2-Audio} Technical Report},
  author={Chu, Yunfei and Xu, Jin and Yang, Qian and Wei, Haojie and Wei, Xipin and Guo, Zhifang and Leng, Yichong and Lv, Yuanjun and He, Jinzheng and Lin, Junyang and Zhou, Chang and Zhou, Jingren},
  journal={arXiv preprint arXiv:2407.10759},
  year={2024},
  url={https://arxiv.org/abs/2407.10759}
}

@inproceedings{shi2023large,
  title={Large Language Models Can Be Easily Distracted by Irrelevant Context},
  author={Shi, Freda and Chen, Xinyun and Misra, Kanishka and Scales, Nathan and Dohan, David and Chi, Ed H. and Sch{\"a}rli, Nathanael and Zhou, Denny},
  booktitle={Proceedings of the 40th International Conference on Machine Learning},
  pages={31210--31227},
  year={2023}
}


\clearpage
\appendix

\section{Dataset Statistics and Evaluation Metrics}
\label{app:datasets}

The dataset splits are summarized in Table~\ref{tab:dataset-splits}, and the class-wise support in the test sets is given in Table~\ref{tab:class-support}.

\begin{table}[!htbp]
\centering

\caption{Dataset splits for IEMOCAP and MELD.}
\label{tab:dataset-splits}

\small
\renewcommand{\arraystretch}{1.5}
\setlength{\tabcolsep}{6pt}

\begin{tabular*}{\columnwidth}{@{\extracolsep{\fill}}lccc@{}}
\toprule

\textbf{Dataset}
& \textbf{Train}
& \textbf{Dev}
& \textbf{Test} \\

\midrule

IEMOCAP & 5,112 & 646 & 1,622 \\
MELD    & 9,989 & 1,109 & 2,610 \\

\bottomrule
\end{tabular*}

\end{table}

\begin{table}[!htbp]
\centering

\caption{Test-set class support on IEMOCAP and MELD.}
\label{tab:class-support}

\small
\renewcommand{\arraystretch}{1.25}
\setlength{\tabcolsep}{5pt}

\begin{tabular*}{\columnwidth}{@{\extracolsep{\fill}}lclc@{}}
\toprule

\multicolumn{2}{c}{\textbf{IEMOCAP}}
& \multicolumn{2}{c}{\textbf{MELD}} \\

\cmidrule(lr){1-2}
\cmidrule(lr){3-4}

\textbf{Class}
& \textbf{Support}
& \textbf{Class}
& \textbf{Support} \\

\midrule

Neutral    & 384 & Neutral  & 1,256 \\
Happy      & 143 & Joy      & 402   \\
Sad        & 245 & Surprise & 281   \\
Angry      & 170 & Anger    & 345   \\
Excited    & 299 & Sadness  & 208   \\
Frustrated & 381 & Disgust  & 68    \\
--         & --  & Fear     & 50    \\

\midrule

\textbf{Total}
& \textbf{1,622}
& \textbf{Total}
& \textbf{2,610} \\

\bottomrule
\end{tabular*}

\end{table}

For emotion class $c$, precision and recall are defined as
\begin{equation}
P_c=\frac{TP_c}{TP_c+FP_c},
\qquad
R_c=\frac{TP_c}{TP_c+FN_c},
\end{equation}
where $TP_c$, $FP_c$, and $FN_c$ denote the numbers of true positives, false positives, and false negatives, respectively.

Class-wise F1 is defined as:
\begin{equation}
F1_c=\frac{2P_cR_c}{P_c+R_c}.
\end{equation}

Accuracy is defined as:
\begin{equation}
\mathrm{Acc.}
=
\frac{\sum_{c=1}^{C}TP_c}{N},
\end{equation}
where $C$ is the number of emotion classes and $N$ is the total number of evaluated utterances.

Weighted F1 is defined as:
\begin{equation}
\mathrm{W\text{-}F1}
=
\sum_{c=1}^{C}
\frac{n_c}{N}F1_c,
\end{equation}
where $n_c$ denotes the number of ground-truth samples in class $c$.

Macro-F1 is defined as:
\begin{equation}
\mathrm{M\text{-}F1}
=
\frac{1}{C}
\sum_{c=1}^{C}F1_c.
\end{equation}


\section{Implementation Details}
\label{app:impl}

The VLM adaptation settings are listed in Table~\ref{tab:vlm-adaptation-settings}.

\begin{table}[!htbp]
\centering

\caption{VLM adaptation settings.}
\label{tab:vlm-adaptation-settings}

\footnotesize
\renewcommand{\arraystretch}{1.35}
\setlength{\tabcolsep}{5pt}

\begin{tabular*}{\columnwidth}{@{\extracolsep{\fill}}ll@{}}
\toprule

\textbf{Setting}
& \textbf{Value} \\

\midrule

VLM backbone & Qwen3-VL-8B-Instruct \\
LoRA target modules & q/k/v/o projections \\
LoRA rank & 16 \\
LoRA alpha & 16 \\
LoRA dropout & 0.05 \\
Optimizer & AdamW \\
Weight decay & 0.01 \\
Learning rate & $2\times10^{-4}$ \\
Epochs & 4 \\
LR schedule & Cosine; 3\% warmup \\

\bottomrule
\end{tabular*}

\end{table}

The dataset-specific hyperparameters of the coupling module are listed in Table~\ref{tab:fusion-settings}.

\begin{table}[!htbp]
\centering

\caption{Coupling training settings.}
\label{tab:fusion-settings}

\small
\renewcommand{\arraystretch}{1.4}
\setlength{\tabcolsep}{6pt}

\begin{tabular*}{\columnwidth}{@{\extracolsep{\fill}}lcc@{}}
\toprule

\textbf{Setting}
& \textbf{IEMOCAP}
& \textbf{MELD} \\

\midrule

Optimizer & AdamW & AdamW \\
Batch size & 128 & 128 \\
Epochs & 30 & 10 \\
Learning rate
& $4\times10^{-5}$
& $8\times10^{-6}$ \\
Dropout & 0.45 & 0.40 \\
LR schedule & Cosine & Cosine \\
InfoNCE temperature & 0.1 & 0.1 \\
$\lambda_{\mathrm{CL}}$ & 0.1 & 0.1 \\
$\lambda_{\mathrm{aux}}$ & 0.2 & 0.1 \\

\bottomrule
\end{tabular*}

\end{table}


\section{Prompt Designs}
\label{app:prompts}

\begin{figure}[H]
\centering

\subfloat[\normalfont Audio-description generation prompt.]{%
\begin{minipage}[t]{1\columnwidth}
    \centering
    \includegraphics[width=\linewidth]
    {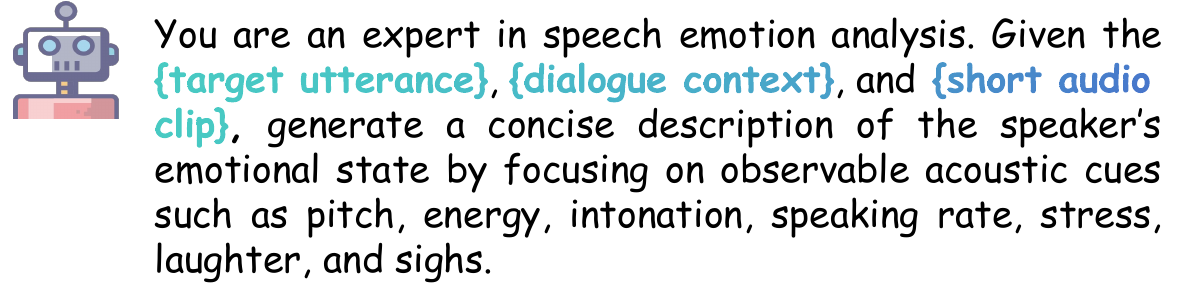}
\end{minipage}}

\par\vspace{4pt}

\subfloat[\normalfont Emotion-supervised VLM adaptation prompt.]{%
\begin{minipage}[t]{1\columnwidth}
    \centering
    \includegraphics[width=\linewidth]
    {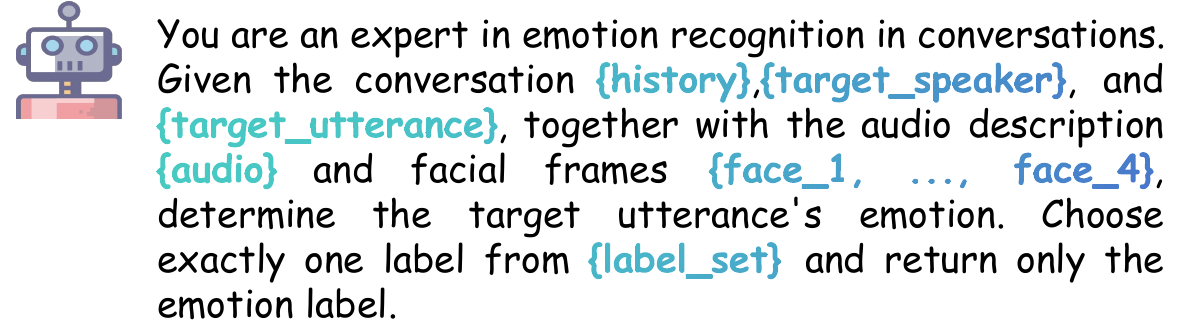}
\end{minipage}}

\caption{
    Prompt designs for \method{}.
}
\label{fig:prompt-design}

\end{figure}

Figure~\ref{fig:prompt-design} presents the prompts for target-utterance audio-description generation and emotion-supervised VLM adaptation.

\section{Learning Objective Details}
\label{app:objective}

\subsection{Cross-Modal Contrastive Learning}

To align the affective semantics of the textual, acoustic, and contextual representations, we add cross-modal contrastive learning.
We use three independent representation-specific MLP projection heads that map the 512-dimensional $t$, $\bar{A}$, and $v$ into a 128-dimensional contrastive space:
\begin{equation}
z_i^t = P_t(t_i), \qquad
z_i^A = P_A(\bar{A}_i), \qquad
z_i^v = P_v(v_i),
\label{eq:contrastive-feature}
\end{equation}
where $P_t$, $P_A$, and $P_v$ denote the projection heads for the textual, acoustic, and contextual representations, respectively.
The projected representations are L2-normalized before the contrastive similarity is computed:
\begin{equation}
\hat{z}_i^m
=
\frac{z_i^m}{\|z_i^m\|_2},
\qquad
m\in\{t,A,v\}.
\label{eq:l2norm}
\end{equation}

For a pair of representation types $p$ and $q$, the directional InfoNCE loss is defined as:
\begin{equation}
\mathcal{L}_{p\rightarrow q}
=
-\frac{1}{B}
\sum_{i=1}^{B}
\log
\frac{
\exp\left(
(\hat{z}_i^p)^\top
\hat{z}_i^q/\tau
\right)
}{
\sum_{j=1}^{B}
\exp\left(
(\hat{z}_i^p)^\top
\hat{z}_j^q/\tau
\right)
},
\label{eq:infonce}
\end{equation}

where $B$ denotes the batch size and $\tau$ is the temperature parameter.
For each pair of representations, we use only one directional InfoNCE term to avoid duplicating the pairwise alignment objective.
Representations of different types from the same target utterance form positive pairs, and representations from other utterances in the mini-batch serve as negatives.
The overall cross-modal contrastive learning objective is:
\begin{equation}
\mathcal{L}_{\mathrm{CL}}
=
\mathcal{L}_{t\rightarrow A}
+
\mathcal{L}_{t\rightarrow v}
+
\mathcal{L}_{A\rightarrow v}.
\label{eq:contrastive-loss}
\end{equation}
The contrastive projection heads are used only to compute the contrastive objective and do not take part in residual coupling, and the original 512-dimensional representations are still used for local cue extraction and coupling.

\subsection{Contextual Representation Auxiliary Loss}

To preserve the emotion-discriminative information of the contextual representation, we add an auxiliary emotion classifier on $v$.
For the $i$-th target utterance, the contextual representation is mapped to an emotion probability distribution:
\begin{equation}
\hat{\mathbf{p}}_i^v
=
\operatorname{Softmax}
\left(
W_{\mathrm{aux}}v_i+b_{\mathrm{aux}}
\right),
\label{eq:aux-prob}
\end{equation}

where $W_{\mathrm{aux}}$ and $b_{\mathrm{aux}}$ are trainable parameters of the auxiliary classifier.
The auxiliary classification loss is defined as:
\begin{equation}
\mathcal{L}_v
=
-\frac{1}{B}
\sum_{i=1}^{B}
\sum_{c=1}^{C}
y_{i,c}
\log
\hat{p}_{i,c}^v.
\label{eq:aux-loss}
\end{equation}
The auxiliary classifier is used only during training and is removed at inference.


\section{Training Procedure}
\label{app:training}

\begin{algorithm}[t]
\caption{Training Procedure of \method{}}
\label{alg:training}

\begin{algorithmic}[1]

\REQUIRE
Dataset splits $\mathcal{D}_{\mathrm{train}}$, $\mathcal{D}_{\mathrm{dev}}$, and $\mathcal{D}_{\mathrm{test}}$;
pretrained VLM parameters $\theta_0$;
LoRA parameters $\Delta\theta$;
coupling parameters $\Phi$

\ENSURE
Trained \method{} model

\STATE
\textbf{VLM adaptation and feature extraction}

\STATE
Initialize LoRA parameters $\Delta\theta$

\FOR{each adaptation epoch}
    \FOR{each mini-batch from the training split}
        \STATE Construct multimodal conditional input $X$
        \STATE Compute $\mathcal{L}_{\mathrm{adapt}}$ using Eq.~\ref{eq:adapt}
        \STATE Update $\Delta\theta$ while keeping $\theta_0$ frozen
    \ENDFOR
\ENDFOR

\STATE
For each data split, extract and cache $t_{\mathrm{raw}}$, $A_{\mathrm{raw}}$, and $v_{\mathrm{raw}}$ using the fixed feature extractors

\STATE
\textbf{Sample-adaptive residual coupling}

\STATE
Initialize coupling parameters $\Phi$

\FOR{each training epoch}
    \FOR{each mini-batch from the training split}
        \STATE Load cached $t_{\mathrm{raw}}$, $A_{\mathrm{raw}}$, and $v_{\mathrm{raw}}$
        \STATE Map the representations into the shared feature space to obtain $t$, $A$, $\bar{A}$, and $v$
        \STATE Extract context-guided acoustic cues $e^a$ using Eq.~\ref{eq:cross-attention}
        \STATE Compute residuals, gating coefficients, and the fused representation $v^*$ using Eq.~\ref{eq:adaptive-refinement}
        \STATE Compute $\mathcal{L}_{\mathrm{cls}}$, $\mathcal{L}_{\mathrm{CL}}$, and $\mathcal{L}_v$
        \STATE Compute $\mathcal{L}_{\mathrm{total}}$ using Eq.~\ref{eq:total-loss}
        \STATE Update $\Phi$ while keeping all feature extractors frozen
    \ENDFOR
\ENDFOR

\RETURN
Trained \method{} model

\end{algorithmic}
\end{algorithm}

Algorithm~\ref{alg:training} presents the training procedure of \method{}, consisting of emotion-supervised VLM adaptation, feature extraction, and sample-adaptive residual coupling.
Lines 1--9 describe the VLM adaptation stage.
The LoRA parameters are initialized and optimized over multiple epochs using the training split.
For each mini-batch, the multimodal input $X$ is constructed from the target utterance, target-speaker visual information, preceding dialogue, and acoustic description.
The adaptation loss is then computed over the ground-truth emotion label tokens, and only the LoRA parameters are updated while the pretrained VLM backbone remains frozen.
After adaptation, Line 10 extracts the textual features $t_{\mathrm{raw}}$, frame-level acoustic features $A_{\mathrm{raw}}$, and continuous contextual representations $v_{\mathrm{raw}}$ using RoBERTa-large, Emotion2vec-large, and the adapted VLM, respectively.
These features are cached offline for subsequent training.
Lines 11--23 describe the sample-adaptive residual coupling stage.
For each mini-batch, the cached features are loaded and mapped into a shared feature space (Lines 15--16).
In Line 17, the contextual representation $v$ serves as the query in multi-head cross-attention to extract context-guided acoustic cues $e^A$ from the frame-level acoustic sequence $A$.
Line 18 computes textual and acoustic residuals relative to $v$ and adjusts their contributions through sample-adaptive gating to obtain the fused representation $v^*$.
Lines 19--20 compute the classification, cross-modal contrastive, and auxiliary classification losses and combine them into the overall training objective.
The coupling parameters and associated classification modules are then optimized while keeping the upstream feature extractors frozen (Line 21).
Finally, Line 24 returns the trained \method{} model.
\section{Additional Experimental Results}
\label{app:additional-results}


\subsection{Effect of Training Objectives}
\label{app:objective-ablation}

This section presents supplementary analyses of training objectives and overall local-cue complementarity.
Beyond the coupling structure, we run supplementary ablations on cross-modal contrastive learning (CL) and the contextual representation auxiliary classification loss (Aux).
We remove CL or Aux alone and then both together, keeping the model architecture, input features, data splits, optimization settings, and random seeds the same across variants.
When both are removed, the model is trained with the final classification loss only.
The results are summarized in Table~\ref{tab:objective-ablation}.

\begin{table}[t]
\centering

\caption{
    Training-objective ablations on IEMOCAP and MELD.
}
\label{tab:objective-ablation}

\footnotesize
\setlength{\tabcolsep}{3pt}
\renewcommand{\arraystretch}{1.08}

\begin{tabular*}{\columnwidth}{@{\extracolsep{\fill}}lccc@{}}
\toprule

\textbf{Variant}
& \textbf{Acc.}
& \textbf{W-F1}
& \textbf{M-F1} \\

\midrule

\multicolumn{4}{l}{\textbf{IEMOCAP}} \\

Full \method{}
& $76.99\pm0.23$
& $77.07\pm0.22$
& $76.41\pm0.26$ \\

w/o CL
& $76.35\pm0.28$
& $76.44\pm0.22$
& $75.84\pm0.39$ \\

w/o Aux
& $76.72\pm0.18$
& $76.78\pm0.09$
& $76.07\pm0.23$ \\

w/o CL and Aux
& $76.04\pm0.36$
& $76.03\pm0.27$
& $75.13\pm0.30$ \\

\midrule

\multicolumn{4}{l}{\textbf{MELD}} \\

Full \method{}
& $70.43\pm0.45$
& $70.07\pm0.37$
& $57.46\pm0.52$ \\

w/o CL
& $69.96\pm0.20$
& $69.56\pm0.20$
& $56.97\pm0.26$ \\

w/o Aux
& $70.19\pm0.45$
& $69.89\pm0.34$
& $57.33\pm0.43$ \\

w/o CL and Aux
& $69.65\pm0.86$
& $69.36\pm0.67$
& $56.54\pm0.99$ \\

\bottomrule
\end{tabular*}

\end{table}

Removing either objective lowers all three overall metrics, and removing both gives the largest drop.
On IEMOCAP, removing both CL and Aux lowers M-F1 from 76.41\% to 75.13\%, a drop of 1.28 points.
On MELD, it lowers M-F1 from 57.46\% to 56.54\%, a drop of 0.92 points.

Removing CL alone has a larger effect than removing Aux alone.
M-F1 falls by 0.57 and 0.49 points on IEMOCAP and MELD without CL, compared with 0.34 and 0.13 points without Aux.
Overall, both auxiliary objectives contribute to performance, with CL showing a larger contribution.


\subsection{Overall Local-Cue Complementarity}
\label{app:complementarity}

The main text examines class-wise local-cue contributions and the standalone discriminative capabilities of textual and acoustic representations.
Here, we further compare overall M-F1 across different local-cue configurations.

\begin{figure}[t]
    \centering

    \begin{minipage}{\linewidth}
        \centering

        \includegraphics[width=\linewidth]
        {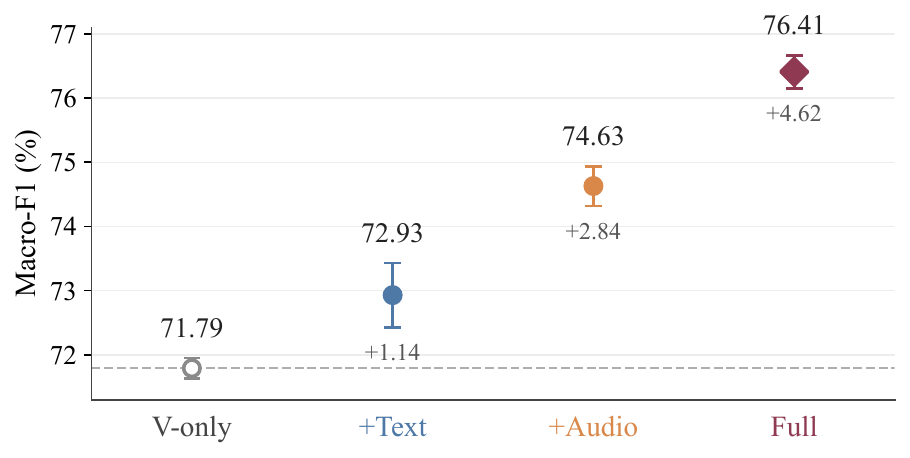}

        {\small (a) IEMOCAP}
    \end{minipage}

    \par\vspace{4pt}

    \begin{minipage}{\linewidth}
        \centering

        \includegraphics[width=\linewidth]
        {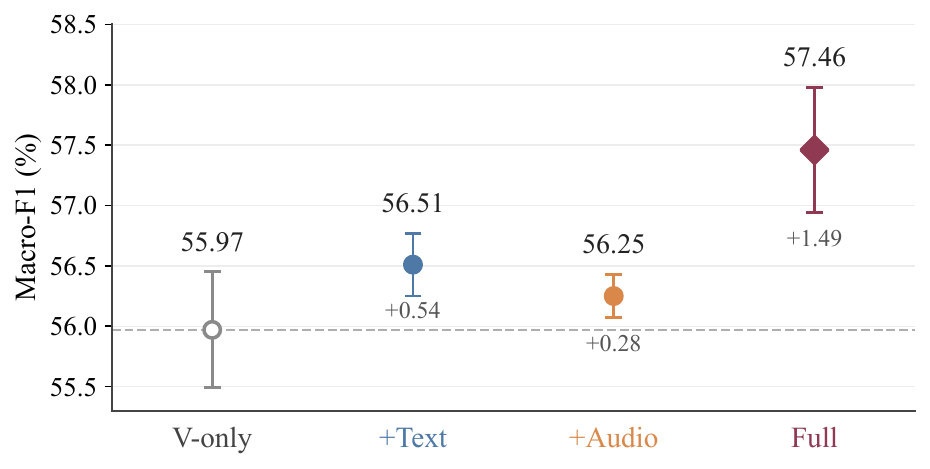}

        {\small (b) MELD}
    \end{minipage}

    \caption{
        Overall complementarity of local textual and acoustic cues on IEMOCAP and MELD.}
        
    \label{fig:overall-complementarity}
\end{figure}

Figure~\ref{fig:overall-complementarity} compares V-only with variants that add textual cues, acoustic cues, or both.
Compared with V-only, Full \method{} improves M-F1 from 71.79 to 76.41 on IEMOCAP and from 55.97 to 57.46 on MELD.
The contributions of individual modalities vary across the two datasets.

\section{Notation Summary}
\label{app:notation}

Table~\ref{tab:notation} summarizes the main mathematical symbols used in the formulations of VLM adaptation, multimodal feature representations, residual coupling, and training objectives in CLEAR.

\begin{table}[t]
\centering
\caption{Summary of the main notation used in CLEAR.}
\label{tab:notation}

\footnotesize
\setlength{\tabcolsep}{4pt}
\renewcommand{\arraystretch}{1.08}

\begin{tabularx}{\columnwidth}{@{}lX@{}}
\toprule
\textbf{Notation} & \textbf{Description} \\
\midrule


\multicolumn{2}{@{}l}{\emph{Input and Feature Representations}} \\

$X$ & Multimodal input to the VLM. \\

$\theta_0,\Delta\theta_{\mathrm{LoRA}}$ & Frozen VLM parameters and trainable LoRA parameters. \\

$L,p$ & Final VLM layer and last valid input-prompt position. \\

$\mathcal{Y}$ & Positions of the ground-truth emotion-label tokens. \\

$\mathbf{v}_{\mathrm{raw}}$ & VLM hidden-state representation (4096 dimensions). \\

$\mathbf{t}_{\mathrm{raw}}$ & Utterance-level textual feature from RoBERTa-large. \\

$\mathbf{A}_{\mathrm{raw}}$ & Frame-level acoustic features from Emotion2vec-large. \\

$\mathbf{v},\mathbf{t}$ & Contextual and textual representations in the shared space. \\

$\mathbf{A},\bar{\mathbf{A}}$ & Frame-level acoustic sequence and its pooled representation. \\

$T,\pi_\ell$ & Number of valid acoustic frames and attention weight for frame $\ell$. \\

\addlinespace[2pt]


\multicolumn{2}{@{}l}{\emph{Residual Coupling and Prediction}} \\

$\mathbf{e}^{A}$ & Context-guided acoustic feature from cross-attention. \\

$\Delta^{t},\Delta^{A}$ & Textual and acoustic residuals relative to $\mathbf{v}$. \\

$\alpha^{t},\alpha^{A}$ & Sample-adaptive gates for the two residuals. \\

$G$ & Learnable residual gating network. \\

$\mathbf{v}^{*}$ & Fused representation after residual coupling and LayerNorm. \\

$f_{\mathrm{cls}},\hat{\mathbf{p}}$ & Final classifier and predicted emotion distribution. \\

$\Phi$ & Trainable parameters of the coupling stage. \\

\addlinespace[2pt]


\multicolumn{2}{@{}l}{\emph{Optimization and Losses}} \\

$B,C$ & Mini-batch size and number of emotion classes. \\

$y_{i,c},\hat{p}_{i,c}$ & Ground-truth class indicator and predicted class probability. \\

$P_t,P_A,P_v$ & Modality-specific contrastive projection heads. \\

$\hat{\mathbf{z}}^{m}_i$ & Normalized contrastive embedding for sample $i$ and modality $m$. \\

$\tau$ & InfoNCE temperature parameter. \\

$\mathcal{L}_{\mathrm{adapt}}$ & Emotion-label loss for VLM adaptation. \\

$\mathcal{L}_{\mathrm{cls}}$ & Final emotion classification loss. \\

$\mathcal{L}_{\mathrm{CL}},\mathcal{L}_{v}$ & Cross-modal contrastive and contextual auxiliary losses. \\

$\lambda_{\mathrm{CL}},\lambda_{\mathrm{aux}}$ & Weights of the two auxiliary objectives. \\

$\mathcal{L}_{\mathrm{total}}$ & Overall coupling-stage training objective. \\

\bottomrule
\end{tabularx}
\end{table}

\end{document}